\documentclass[10pt,journal,compsoc]{IEEEtran}
\usepackage{amsmath,amsfonts}
\usepackage{algorithmic}
\usepackage{algorithm}
\usepackage{array}
\usepackage[caption=false,font=normalsize,labelfont=sf,textfont=sf]{subfig}
\usepackage{textcomp}
\usepackage{stfloats}
\usepackage{url}
\usepackage{verbatim}
\usepackage{graphicx}
\usepackage{cite}
\usepackage{bm}

\usepackage{mathrsfs}
\usepackage{amssymb}
\usepackage{textcomp}
\usepackage{amsmath}
\usepackage{amsthm} 
\newtheorem{theorem}{Theorem}

\newtheorem{lemma}{Lemma}
\newtheorem{definition}{Definition}
\usepackage{booktabs}
\usepackage{multirow}
\usepackage{algorithm}
\usepackage{algorithmic}
\usepackage{diagbox}
\usepackage{xcolor}
\begin{document}

\title{TRUST-FS: Tensorized Reliable Unsupervised Multi-View Feature Selection for Incomplete Data}

\author{Minghui Lu, Yanyong Huang,  Minbo Ma, Jinyuan Chang, Dongjie Wang, Xiuwen Yi,~and~Tianrui Li,~\IEEEmembership{Senior Member,~IEEE}
\thanks{Minghui~Lu, Yanyong~Huang,~and~Jinyuan~Chang are with the Joint Laboratory of Data Science and Business Intelligence, School of Statistics and Data Science, Southwestern University of Finance and Economics, Chengdu 611130, China (e-mail: lu.m.h@foxmail.com; huangyy@swufe.edu.cn;changjinyuan@swufe.edu.cn), Yanyong Huang is the corresponding author;}

\thanks{Minbo Ma is with the Insitute for Carbon Neutrality, Tsinghua University, BeiJing 100084, China (e-mail: minbo46.ma@gmail.com);}

\thanks{Dongjie Wang is with the Department of Electrical Engineering and Computer Science, University of Kansas, Lawrence, KS 66045, USA (e-mail: wangdongjie@ku.edu);}

\thanks{Xiuwen~Yi is with the JD Intelligent Cities Research and JD Intelligent Cities Business Unit, Beijing 100176, China (e-mail: xiuwenyi@foxmail.com);}

\thanks{Tianrui Li is with the School of Computing and Artificial Intelligence, Southwest Jiaotong University, Chengdu 611756, China (e-mail: trli@swjtu.edu.cn).}}

\markboth{
	Journal of IEEE Transactions on Knowledge and Data Engineering,~Vol.~14, No.~8, August~2021  
}
{
	Lu \MakeLowercase{\textit{et al.}}: TRUST-FS: Tensorized Reliable Unsupervised Multi-View Feature Selection for Incomplete Data 
}

\IEEEpubid{0000--0000/00\$00.00~\copyright~2021 IEEE}


\maketitle

\begin{abstract}
Multi-view unsupervised feature selection (MUFS), which selects informative features from multi-view unlabeled data, has attracted increasing research interest in recent years. Although great efforts have been devoted to MUFS, several challenges remain: 1) existing methods for incomplete multi-view data are limited to handling missing views and are unable to address the more general scenario of missing variables, where some features have missing values in certain views; 2) most methods address incomplete data by first imputing missing values and then performing feature selection, treating these two processes independently and overlooking their interactions; 3) missing data can result in an inaccurate similarity graph, which reduces the performance of feature selection. To solve this dilemma, we propose a novel MUFS method for incomplete multi-view data with missing variables, termed Tensorized Reliable UnSupervised mulTi-view Feature Selection (TRUST-FS). TRUST-FS introduces a new adaptive-weighted CP decomposition that simultaneously performs feature selection, missing-variable imputation, and view weight learning within a unified tensor factorization framework. By utilizing Subjective Logic to acquire trustworthy cross-view similarity information, TRUST-FS facilitates learning a reliable similarity graph, which subsequently guides feature selection and imputation. Comprehensive experimental results demonstrate the effectiveness and superiority of our method over state-of-the-art methods.
\end{abstract}


\section{Introduction}
\begin{figure}[!htbp]
	\centering
	\includegraphics[width=\columnwidth]{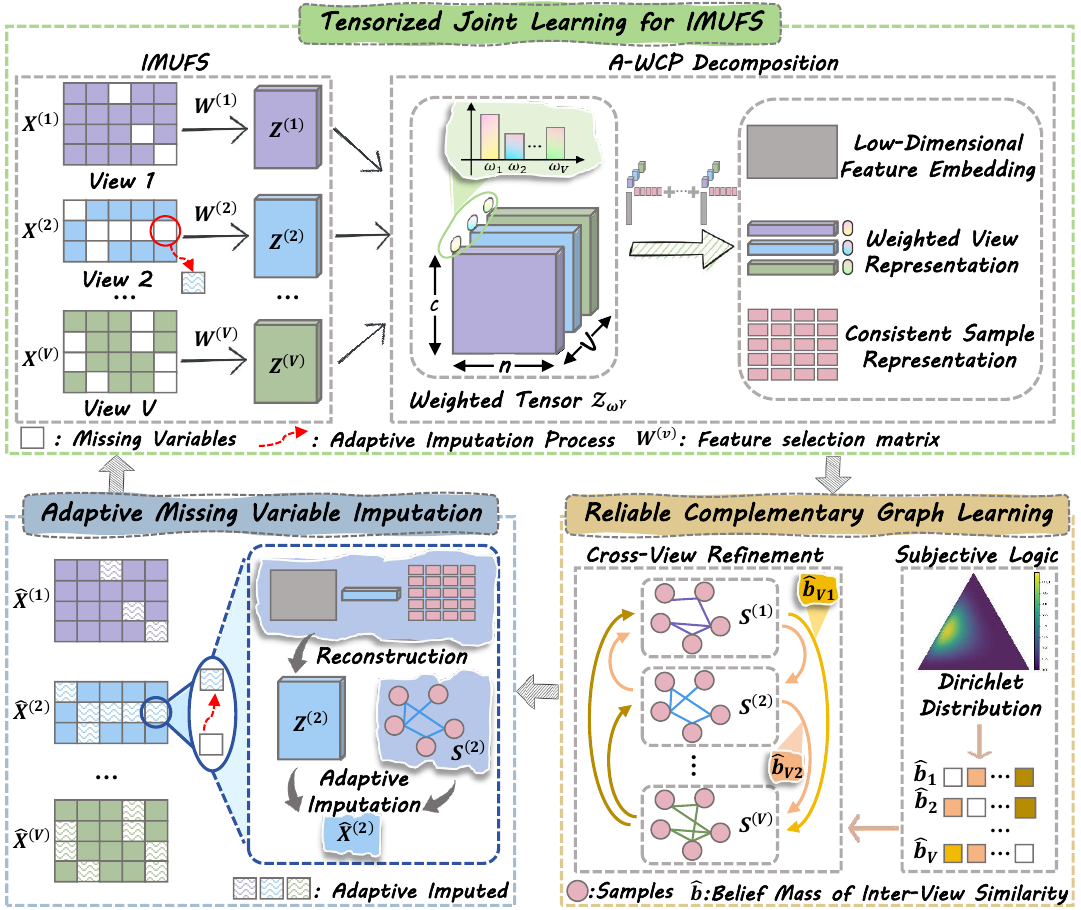} 
	\caption{The framework of the proposed TRUST-FS.}
	\label{Framework}
\end{figure}

\IEEEPARstart{T}{he} rapid development of information technology has led to the widespread presence of multi-view data, in which heterogeneous features from multiple perspectives are used to describe the same sample~\cite{multi-view-survey-2,multi-view-learning}. Multi-view data are often high-dimensional and unlabeled, as their features come from multiple sources and annotation requires significant manual effort~\cite{FS-review, unlabeled}. Multi-view unsupervised feature selection (MUFS), which identifies a subset of the most informative features, has emerged as an effective approach for dimensionality reduction in unlabeled multi-view data~\cite{MFS-review,SCMvFS,zhang2024efficient}.

Existing MUFS methods can generally be divided into two categories. The first category concatenates features from all views into a single set and then applies conventional single-view feature selection methods, such as LS~\cite{LS} and GAWFS~\cite{GAWFS}. In contrast to the first category, which uses feature concatenation, the second category of methods directly selects features from multi-view data by considering the correlations among different views~\cite{CFSMO,WLTL,UKMFS}. Typically, CFSMO \cite{CFSMO} leverages high-order neighbor information to construct similarity graphs and learns a unified latent representation to facilitate feature selection. WLTL~\cite{WLTL} employs a weighted tensor nuclear norm to capture high-order dependencies among views for feature selection.

Although previous methods have shown promising results in feature selection, most rely on the assumption that all samples are completely observed across all views. This assumption may not always be valid in real-world scenarios. For example, in video monitoring, the same scene is often captured simultaneously by multiple cameras from different angles.  While some cameras may be functioning, occlusion can cause the corresponding variables in those views to be missing~\cite{missing-value1}. MUFS methods that assume complete views typically require the imputation of missing values before performing feature selection on incomplete multi-view data. Nonetheless, treating imputation and feature selection as separate processes overlooks the interaction between them, which can result in suboptimal feature selection performance. Only a few studies have explored incomplete multi-view unsupervised feature selection (IMUFS) for handling incomplete data~\cite{C2IMUFS,UNIFIER}. C$^2$IMUFS~\cite{C2IMUFS} utilizes both complementary and consensus information from multiple views to improve unsupervised feature selection for incomplete multi-view data. However, these methods can only handle the view-missing scenario, where entire views are absent for certain samples, and cannot address the more general variable-missing scenario, where some features of certain samples are missing in some views. 
Moreover, most graph-based multi-view feature selection methods have difficulty handling incomplete multi-view data, especially when the missing rate is relatively high. The resulting similarity graphs are often unreliable, which in turn reduces the effectiveness of feature selection.

To address the aforementioned problems, we propose a novel MUFS method for incomplete multi-view data with missing variables, referred to as Tensorized Reliable UnSupervised mulTi-view Feature Selection (TRUST-FS). Specifically, we first propose a tensorized joint learning framework that seamlessly integrates multi-view feature selection, missing variable imputation, and view weight learning through a novel method called Adaptive-Weighted CP decomposition (A-WCP). A-WCP effectively captures both cross-view consistency and view-specific information, thereby facilitating feature selection and imputation. Then, we employ Subjective Logic to assess the reliability of inter-view similarities using belief mass. This belief mass is subsequently utilized to refine each view’s similarity graph by incorporating trustworthy cross-view information, thus enhancing graph reliability and offering more effective guidance for feature selection. Fig.~\ref{Framework} illustrates the overall framework of the proposed TRUST-FS.
The main contributions of this paper are summarized as follows:

\begin{itemize}
	\item To the best of our knowledge, this is the first work to address unsupervised feature selection for incomplete multi-view data with missing variables. It seamlessly integrates multi-view feature selection, missing value imputation, and view-weight learning into a unified learning framework, where these components jointly contribute to improving feature selection performance.
	
	\item TRUST-FS employs the proposed adaptive-weighted CP decomposition to capture both cross-view consistency and view-specific information, thereby facilitating missing data imputation and discriminative feature selection. Additionally, subjective logic is incorporated to enhance the reliability of similarity graphs by leveraging trustworthy cross-view information.
	
	\item An effective  alternative optimization algorithm is designed to solve the proposed TRUST-FS, and extensive experimental results convincingly demonstrate the superiority of TRUST-FS over state-of-the-art methods.
\end{itemize}

\section{Related Work}
Multi-view feature selection methods are typically divided into two categories. The first category merges the features from all views and then applies conventional single-view-based feature selection methods. NDFS \cite{NDFS} is a representative single-view-based feature selection method that employs spectral clustering to identify clusters and leverages the cluster structure to guide feature selection. GAWFS combines adaptive graph learning and nonnegative matrix factorization to obtain a sparse feature-weighting matrix. However, these approaches combine features from all views into a single feature space, overlooking inter-view dependencies and potentially compromising the quality of the selected features. In contrast, methods in the second category perform feature selection by capturing latent relationships across views, rather than relying solely on feature concatenation. Typical methods include the following: CFSMO \cite{CFSMO} exploits high-order neighborhood information to improve graph construction and maps multi-view data into a shared latent space, thereby capturing feature-level complementarity to guide feature selection; UKMFS \cite{UKMFS} constructs a consistent similarity graph by adaptively fusing view-specific graphs and employs binary hashing, thus facilitating unsupervised feature selection.

While the methods discussed above are effective for multi-view data, they typically assume that every sample is fully present in all views. In reality, however, some samples may be partially absent from certain views, a situation known as the variable-missing scenario. The more extreme case, where entire views are missing, is referred to as the view-missing scenario. Recent research on incomplete multi-view feature selection has mainly focused on the view-missing scenario. A representative method, C$^2$IMUFS \cite{C2IMUFS}, utilizes complementary information from different views to reconstruct complete similarity matrices and learns a consensus clustering indicator matrix. These components jointly guide feature selection while preserving local geometric structures. However, C$^2$IMUFS fails to account for the intrinsic relationship between feature selection and data imputation, which may lead to inconsistencies during the pre-imputation process. To address this limitation, several methods employ joint learning frameworks that simultaneously perform feature selection and missing data imputation \cite{UNIFIER,TIME-FS,TERUIMUFS}. UNIFIER \cite{UNIFIER} constructs similarity-based graphs in both sample and feature spaces to iteratively guide view recovery during feature selection, and uses half-quadratic minimization to mitigate the impact of outliers. TIME-FS \cite{TIME-FS} applies CP decomposition to a third-order tensor of low-dimensional view representations to learn a consistent anchor graph and an adaptive anchor set, both of which jointly guide feature selection and missing view imputation. TERUIMUFS \cite{TERUIMUFS} uses robust self-representation learning with a low-rank tensor constraint to recover missing samples, while simultaneously learning a consensus graph representation to guide feature selection.

Although these methods are effective in view-missing scenarios, they still have several limitations. First, when applied to the variable-missing situation, they may not fully utilize partially observed samples, which may cause valuable information to be overlooked. Second, under relatively high missing rates, graph-based methods do not explicitly account for the reliability of the learned similarity graphs, which can undermine the performance of feature selection.

\section{The Propose Method}\label{sec:proposed method}
\subsection{Notations and Problem Definition}
Throughout this paper, tensors are written as bold calligraphic letters (e.g., $\bm{\mathcal{A}}$), while matrices are denoted by bold uppercase letters  (e.g., $\bm{A}$). Given any matrix $\bm{A}$, $\bm{A}_{i.}$, $\bm{A}_{.j}$, and $A_{ij}$ represent its $i$-th row, $j$-th column, and $(i,j)$-th entry, respectively. The $\ell_{2,1}$-norm of $\bm{A}$ is defined as $\|\bm{A}\|_{2,1} = \sum_i \sqrt{\sum_j A_{ij}^2}$.  The Khatri-Rao product of two matrices $\bm{A}\in\mathbb{R}^{n_1\times r}$ and $\bm{B}\in\mathbb{R}^{n_2\times r}$ is defined by $\bm{A}\odot\bm{B}=[\bm{A}_{.1}\otimes\bm{B}_{.1},\ldots,\bm{A}_{.r}\otimes\bm{B}_{.r}]\in\mathbb{R}^{n_{1}n_{2}\times r}$, where $\otimes$ denotes the Kronecker product. For any third-order tensor $\bm{\mathcal{Z}}\in\mathbb{R}^{n_1\times n_2\times n_3}$, the CP decomposition approximates $\bm{\mathcal{Z}}$ as a sum of $r$ rank-one tensors, written as $\bm{\mathcal{Z}}\approx [\![\bm{A},\bm{B},\bm{C}]\!]\equiv\sum_{i=1}^{r}(\bm{A}_{.i}\circ \bm{B}_{.i}\circ \bm{C}_{.i})$, where $\circ$ represents the vector outer product, and $\bm{A} \in \mathbb{R}^{n_1 \times r}$, $\bm{B} \in \mathbb{R}^{n_2 \times r}$, and $\bm{C} \in \mathbb{R}^{n_3 \times r}$ are the factor matrices~\cite{CP-D}.

To clearly define the problem, consider an incomplete multi-view dataset  \( \bm{\mathcal{X}} = \{\bm{X}^{(v)} \in \mathbb{R}^{d_v \times n}\}_{v=1}^V \) with $V$ views, where \( \bm{X}^{(v)} \) denotes the data matrix of the $v$-th view, consisting of \( n \) samples and \( d_v \) features.  We introduce an indicator matrix $\bm{M}^{(v)} \in \{0,1\}^{d_v \times n}$ to describe missing data in $\bm{X}^{(v)}$. Specifically, for the $j$-th sample, \(M^{(v)}_{ij}=0\) if its $i$-th feature is missing in the $v$-th view; otherwise, \(M^{(v)}_{ij}=1\). If some features of certain samples in some views have missing values, this condition is referred to as variable missing. View missing, on the other hand, is a specific case of variable missing, where all features in certain views are missing for some samples. In this study, we aim to identify the $l$ most informative features from the incomplete multi-view dataset $\bm{\mathcal{X}}$, focusing on feature selection under the variable missing condition.

\subsection{Tensorized Joint Learning for IMUFS}

Although many MUFS methods have been proposed, they typically handle incomplete multi-view data with missing variables using a two-step approach: first imputing the missing variable with predefined values, followed by feature selection. This sequential strategy treats feature selection and missing variable imputation as independent processes, neglecting the potential interaction between them. In addition, only a limited number of IMUFS methods have been developed for feature selection on incomplete multi-view data~\cite{TIME-FS}. However, these methods are restricted to addressing the view-missing scenario and fail to tackle the more general case, namely the variable-missing scenario. 

To address the aforementioned issues, we propose a tensorized joint learning framework that seamlessly integrates multi-view feature selection and missing variable imputation. The process begins by projecting the multi-view data into a low-dimensional space and stacking the resulting data matrices to construct a third-order tensor. CP tensor decomposition\cite{CP-D,CP} is then employed to capture cross-view consistency and view-specific information, which collectively guide feature selection and the adaptive imputation of missing data. The proposed framework can be formulated as follows:
\begin{equation}\label{adaptive imputation1}
	\begin{aligned}
		&\min _{\bm{W}^{(v)},\bm{\hat{X}}^{(v)},\atop \bm{A},\bm{H},\bm{P}} \|\bm{\mathcal{Z}}-[\![\bm{A},\bm{H},\bm{P}]\!]\|_{F}^2+\lambda\sum_{v=1}^V\|\bm{W}^{(v)}\|_{2,1}\\
		&{s.t.}~ \bm{\mathcal{Z}}=\Phi(\bm{W}^{(1)^\mathrm{T}}\bm{\hat{X}}^{(1)},\ldots,\bm{W}^{(V)^\mathrm{T}}\bm{\hat{X}}^{(V)}),\bm{\hat{X}}^{(v)}\geq0,\\
		&~~~~~~\bm{M}^{(v)}\circledast\bm{\hat{X}}^{(v)}=\bm{M}^{(v)}\circledast\bm{X}^{(v)},\bm{W}^{(v)},\bm{A},\bm{P},\bm{H}\geq0,\\
	\end{aligned}
\end{equation}
where $\bm{\bm{\mathcal{Z}}}\in\mathbb{R}^{c\times n\times V}$ denotes the third-order tensor formed by stacking the $V$ projected low-dimensional representations $\{\bm{W}^{(v)^\mathrm{T}}\bm{\hat{X}}^{(v)}\}_{v=1}^V$, $\bm{\hat{X}}^{(v)} \in \mathbb{R}^{d_v \times n}$ is an imputed data matrix,  $\bm{W}^{(v)} \in \mathbb{R}^{d_v \times c}$ represents  the feature selection matrix of the $v$-th view, and $\lambda$ serves as  a regularization parameter. In Eq.~(\ref{adaptive imputation1}),  the first term employs CP decomposition to factorize  $\bm{\mathcal{Z}}$ into three factor matrices: $\bm{A} \in \mathbb{R}^{c \times r}$, representing the low-dimensional feature embedding; $\bm{H} \in \mathbb{R}^{n \times r}$, capturing the consistent sample representation across views; and $\bm{P} \in \mathbb{R}^{V \times r}$, characterizing the view-specific representation for each individual view. The second term in Eq.~(\ref{adaptive imputation1}) imposes the $\ell_{2,1}$-norm on the feature selection matrix $\bm{W}^{(v)}$ to enforce row sparsity, thereby facilitating the selection of informative features. Furthermore, the constraint $\bm{M}^{(v)} \circledast \bm{\hat{X}}^{(v)} = \bm{M}^{(v)} \circledast \bm{X}^{(v)}$ ($\circledast$ denotes the Hadamard product) in Eq.~(\ref{adaptive imputation1}) is imposed to ensure that the available entries in the original matrix $\bm{X}^{(v)}$ are identical to those in $\bm{\hat{X}}^{(v)}$. Simultaneously, $\bm{\hat{X}}^{(v)}$ is treated as an optimization variable, enabling the adaptive imputation of missing values in $\bm{X}^{(v)}$.

As different views contribute unequally to the multi-view feature selection task, assigning appropriate view weights is beneficial for improving feature selection performance~\cite{UNIFIER}. The standard CP decomposition in Eq.~(\ref{adaptive imputation1}) fails to account for view importance. To overcome this limitation, we propose an Adaptive-Weighted CP decomposition (A-WCP), which extends the standard CP decomposition by incorporating a view-weighted strategy to better capture view-specific contributions. Applying A-WCP to the aforementioned framework in Eq.~(\ref{adaptive imputation1}) leads to the following formulation:
\begin{equation}\label{adaptive imputation4}
	\begin{aligned}
		&\min _{\bm{W}^{(v)},\bm{\hat{X}}^{(v)},\atop \bm{A},\bm{H},\bm{P},\bm{\omega}}\!\! \|\bm{\bm{\mathcal{Z}}}_{\bm{\omega},{\gamma}}-\![\![\bm{A},\bm{H},\bm{P}_{\bm{\omega},{\gamma}}]\!]\|_{F}^2\!+\!\lambda\sum_{v=1}^V\omega_v^{\gamma}\|\bm{W}^{(v)}\!\|_{2,1}\\
		&{s.t.}~\bm{\bm{\mathcal{Z}}}_{\bm{\omega},{\gamma}}=\Phi(\omega_1^{\frac{\gamma}{2}}\bm{W}^{(1)^\mathrm{T}}\bm{\hat{X}}^{(1)},\ldots,\omega_V^{\frac{\gamma}{2}}\bm{W}^{(V)^\mathrm{T}}\bm{\hat{X}}^{(V)}),\\
		&~~~~~~\bm{P}_{\bm{\omega},\gamma} =[\omega_1^{\frac{\gamma}{2}}\bm{P}_{1.};\ldots;\omega_V^{\frac{\gamma}{2}}\bm{P}_{V.}],\bm{A},\bm{P},\bm{H},\bm{\omega}\!\geq0,\bm{\omega}\bm{1}=1,\\
		&~~~~~~\bm{M}^{(v)}\!\!\circledast\!\bm{\hat{X}}^{(v)}\!\!=\!\bm{M}^{(v)}\!\!\circledast\!\bm{X}^{(v)}\!\!,\bm{W}^{(v)}\!,\!\bm{\hat{X}}^{(v)}\!\!\geq\!0,v=1,\ldots,\!V\!,\\
	\end{aligned}
\end{equation}
where $\bm{\omega}=[\omega_1,\cdots,\omega_V]$ denotes the weights of the $V$ views,  and $\gamma$ is a regularization parameter. In Eq.~(\ref{adaptive imputation4}), the first term corresponds to the proposed A-WCP, where $\bm{\bm{\mathcal{Z}}}_{\bm{\omega},{\gamma}}=\Phi(\omega_1^{\frac{\gamma}{2}}\bm{Z}^{(1)},\cdots,\omega_V^{\frac{\gamma}{2}}\bm{Z}^{(V)})$ denotes the view-weighted tensor, and $\bm{P}_{\bm{\omega},{\gamma}}=[\omega_1^{\frac{\gamma}{2}}\bm{P}_{1.};\cdots;\omega_V^{\frac{\gamma}{2}}\bm{P}_{V.}]$ represents the weighted view-specific representation. A-WCP iteratively learns adaptive view weights within the tensor framework, allowing the model to automatically emphasize informative views while downweighting less relevant ones. At the same time, it leverages both view complementarity and cross-view consistency to enhance the quality of learned representations, including view-specific representations (i.e., $\bm{W}^{(v)\mathrm{T}}\bm{\hat{X}}^{(v)}$), consistent sample representation (i.e., $\bm{H}$), view-level representation (i.e., $\bm{P}$) and low-dimensional feature embedding (i.e., $\bm{A}$). By replacing the standard CP decomposition with A-WCP, the resulting tensorized joint learning framework integrates adaptive view weighting, multi-view feature selection, and missing-value imputation in a synergistic way, further improving both the quality of representations and the effectiveness of feature selection.

\subsection{Reliable Complementary Graph Learning}
Previous studies have demonstrated that constructing similarity-induced graphs can effectively preserve sample similarity relationships and improve feature selection performance~\cite{CFSMO,UKMFS}. Given that multi-view data consist of multiple views representing the same samples, constructing similarity-induced graphs is an effective way to ensure that samples with high similarity in the high-dimensional space maintain their similarity in the unified low-dimensional latent space. Additionally, for samples with missing values, their intrinsic similarity should be preserved during imputation. To this end, we can formulate it as follows:
\begin{equation}\label{confidence1}
	\begin{aligned}
		\min _{\bm{S}^{(v)}}& \sum_{v=1}^{V}\{Tr(\bm{H}^{\mathrm{T}}\bm{L}^{(v)}\bm{H})\!+\!\frac{1}{2}\!\sum_{i,j=1}^n\!\|\bm{\hat{X}}^{(v)}_{.i}-\bm{\hat{X}}^{(v)}_{.j}\|_{2}^{2}S^{(v)}_{ij}\}\\
		{s.t.}~&{S}^{(v)}_{ii}=0,\bm{S}^{(v)\mathrm{T}}\bm{1}=\bm{1},\bm{S}^{(v)}\geq0,v=1,\ldots,V,
	\end{aligned}
\end{equation}
where $\bm{S}^{(v)}\in\mathbb{R}^{n\times n}$ denotes the similarity graph for the $v$-th view, \(\bm{L}^{(v)}=\bm{D}^{(v)}-(\bm{S}^{(v)}+\bm{S}^{(v)\mathrm{T}})/2\) is its corresponding Laplacian matrix, and \(\bm{D}^{(v)}\) is a diagonal matrix with entries $D^{(v)}_{ii}=\sum^{n}_{j=1}(S^{(v)}_{ij}+S^{(v)}_{ji})/2$. In Eq.~(\ref{confidence1}), the learned similarity graph captures the local structure of the data, aiding in the improvement of feature selection performance. Due to the presence of missing data, especially when the missing rate is high, the learned similarity graph fails to accurately capture sample relationships, which in turn adversely affects the performance of feature selection. Building on the complementary relationships among different views in multi-view data, we aim to utilize the similarity information from other views to refine the similarity graph, as described below.
\begin{equation}\label{confidence1N}
	\begin{aligned}
		\min _{\bm{S}^{(v)}}& \sum_{v=1}^{V}\{Tr(\bm{H}^{\mathrm{T}}\bm{L}^{(v)}\bm{H})\!+\!\frac{1}{2}\!\sum_{i,j=1}^n\!\|\bm{\hat{X}}^{(v)}_{.i}-\bm{\hat{X}}^{(v)}_{.j}\|_{2}^{2}S^{(v)}_{ij}\}\\
		& +\sum_{v=1}^{V}\|\bm{S}^{(v)}\!-\!\sum_{k=1,k\neq v}^{V}b_{vk}\bm{S}^{(k)}\|_{F}^{2}\\
		{s.t.}~&S^{(v)}_{ii}=0,\bm{S}^{(v)\mathrm{T}}\bm{1}=\bm{1},\bm{S}^{(v)}\geq0,v=1,\ldots,V,
	\end{aligned}
\end{equation}
where $b_{vk}$ represents the similarity between the $v$-th and $k$-th views, with higher values indicating greater similarity. A straightforward way to compute $b_{vk}$ is using the view-specific representation $\bm{P}$, defined as ${b}_{vk}=({\bm{P}_{v.}\bm{P}_{k.}^{\mathrm{T}}})/{\sqrt{r}}~(v\neq k)$, where $\bm{P}_{v.}$ and $\bm{P}_{k.}$ denote the specific representations of the $v$-th and $k$-th views, respectively. Dividing by $\sqrt{r}$ normalizes the vector length, ensuring that the result is independent of its dimensionality. In Eq.~(\ref{confidence1N}), the third term refines the similarity graph by integrating structural information from highly similar graphs of other views, while reducing the influence of less similar ones. 

Due to the presence of missing values in the incomplete multi-view scenario, they can result in unreliable similarity calculations between views, ultimately compromising the accuracy of the learned similarity graph. To address this limitation, we leverage Subjective Logic (SL)~\cite{SL1,SL2} and the Dirichlet distribution~\cite{dirichlet1,dirichlet2} to enhance the reliability of similarity calculations between views. Specifically, for the $v$-th view, we use the view-specific representation $\bm{P}$ to calculate its similarity to all other views, thereby forming the evidence vector \(\bm{e}_{v}=[e_{vk}]_{k=1,k\neq v}^{V}=[b_{vk}]_{k=1,k\neq v}^{V}\). Within this vector, $e_{vk}$ represents the degree of support (i.e., evidence) for the similarity between view \( v \) and view \( k \). Then, the concentration parameter $\boldsymbol{\alpha}_v$ of the Dirichlet distribution are induced by $\bm{e}_{v}$, i.e., \( \boldsymbol{\alpha}_v = \bm{e}_v + \bm{1} \), where \( \bm{1} \in \mathbb{R}^{V-1} \) denotes an all-ones vector. Based on Subjective Logic, the belief mass vector \(\hat{\bm{b}}_{v}=[\hat{b}_{v1},\ldots,\hat{b}_{v,v-1},\hat{b}_{v,v+1},\ldots,\hat{b}_{vV}]\) and the uncertainty $u_v$ are given by:
\begin{equation}\label{be2}
	\begin{aligned}
		\hat{b}_{vk} = \frac{\alpha_{vk} - 1}{T_v} (k\neq v),~\text{and}~u_v = \frac{V-1}{T_v},
	\end{aligned}
\end{equation}
where $T_v = \sum_{k=1,k\neq v}^{V} \alpha_{vk}$ denotes the Dirichlet strength, indicating the total amount of evidential support. As shown in Eq.~(\ref{be2}), a greater amount of evidence supporting the similarity between the \(v\)-th and \(k\)-th  views yields a higher belief mass, whereas insufficient evidence increases epistemic uncertainty.   The resulting belief mass \( \hat{b}_{vk} \) offers an uncertainty-aware measure of inter-view similarity,  reflecting both evidence sufficiency and epistemic uncertainty within the subjective logic framework. Accordingly, we use \( \hat{b}_{vk} \) in Eq.~(\ref{be2}) to compute the similarity between the $v$-th  and the $k$-th views. This enables us to obtain more reliable similarity graphs from other views, thereby refining the similarity graph by leveraging trustworthy cross-view information.

By combining Eqs. (\ref{adaptive imputation4}), (\ref{confidence1N}) and  (\ref{be2}), we can obtain the final objective of TRUST-FS as follows:
\begin{equation}\label{final_obj}
	\begin{aligned}
		&\min _{\bm{\Psi}} \|\bm{\bm{\mathcal{Z}}}_{\bm{\omega},{\gamma}}-[\![\bm{A},\bm{H},\bm{P}_{\bm{\omega},{\gamma}}]\!]\|_{F}^2+\lambda\sum_{v=1}^V\omega_v^{\gamma}\|\bm{W}^{(v)}\|_{2,1}\\
		&+\tau\sum_{v=1}^{V}\{\frac{1}{2}\sum_{i,j=1}^n\|\bm{\hat{X}}^{(v)}_{.i}-\bm{\hat{X}}^{(v)}_{.j}\|_2^2S^{(v)}_{ij}+\operatorname{Tr}(\bm{H}^\mathrm{T}\bm{L}^{(v)}\bm{H})\\
		&+\|\bm{S}^{(v)}-\sum_{k=1,k\neq v}^{V}\hat{b}_{vk}\bm{S}^{(k)}\|_F^2\} \\
		&{s.t.}~\bm{\bm{\mathcal{Z}}}_{\bm{\omega},{\gamma}}=\Phi(\omega_1^{\frac{\gamma}{2}}\bm{W}^{(1)^\mathrm{T}}\bm{\hat{X}}^{(1)},\ldots,\omega_V^{\frac{\gamma}{2}}\bm{W}^{(V)^\mathrm{T}}\bm{\hat{X}}^{(V)}),\\
		&\bm{P}_{\bm{\omega},\gamma} =[\omega_1^{\frac{\gamma}{2}}\bm{P}_{1.};\ldots;\omega_V^{\frac{\gamma}{2}}\bm{P}_{V.}],\bm{A},\bm{P},\bm{H},\bm{\omega}\!\geq0,\bm{\omega}\bm{1}=1,\\
		&\bm{M}^{(v)}\!\circledast\!\bm{\hat{X}}^{(v)}\!=\bm{M}^{(v)}\!\circledast\!\bm{X}^{(v)}\!\!,\bm{W}^{(v)}\!,\bm{\hat{X}}^{(v)}\!,\bm{S}^{(v)}\!\geq\!0,S^{(v)}_{ii}\!\!=\!0,\\ &\bm{S}^{(v)\mathrm{T}}\bm{1}\!=\!\bm{1},v\!=\!1,\ldots,V,
	\end{aligned}
\end{equation}
where \( \bm{\Psi} = \{ \{ \bm{W}^{(v)}, \bm{\hat{X}}^{(v)}, \bm{S}^{(v)}, \bm{\hat{b}}_{v} \}_{v=1}^V, \bm{A}, \bm{H}, \bm{P},\bm{\omega} \} \), and  \( \tau \) is a trade-off parameter. On the one hand, Eq.~(\ref{final_obj}) builds upon our proposed adaptive-weighted CP decomposition, which seamlessly integrates multi-view feature selection, missing value imputation, and view weight learning into a unified learning framework, allowing these components to mutually enhance one another. On the other hand, it uses trustworthy cross-view information to facilitate reliable similarity graph learning, thereby effectively improving feature selection.

\section{Optimization and Analyses}\label{sec:optimization}
Since the objective function in Eq.~(\ref{final_obj}) is jointly non-convex with respect to all variables, we propose an alternative iterative optimization algorithm to solve this problem.

\noindent\textbf{Update $\bm{\hat{X}}^{(v)}$: }
When fixing other variables, the optimization subproblem for \( \bm{\hat{X}}^{(v)} \)  can be reformulated as follows, based on the definition of CP decomposition~\cite{CP}:
\begin{equation}\label{x3}
	\begin{aligned}
		\min_{\bm{\hat{X}}^{(v)}} ~& \omega_v^{\gamma}\|\bm{W}^{(v)\mathrm{T}}\bm{\hat{X}}^{(v)}-\bm{\Gamma}^{(v)}\|_{F}^2+\tau\operatorname{Tr}(\bm{\hat{X}}^{(v)}\bm{L}^{(v)}\bm{\hat{X}}^{(v)\mathrm{T}})\\
		s.t.~ &\bm{M}^{(v)}\circledast\bm{\hat{X}}^{(v)}=\bm{M}^{(v)}\circledast\bm{X}^{(v)},\bm{\hat{X}}^{(v)}\geq0,
	\end{aligned}
\end{equation}
where $\bm{\Gamma}^{(v)}=\bm{A}diag(\bm{P}_{v.})\bm{H}^\mathrm{T}$. Given the constraint \(\bm{\hat{X}}^{(v)}\geq0\), we introduce a Lagrange multiplier and apply the Karush-Kuhn-Tucker (KKT) conditions~\cite{KKT} to derive the update rule for \( \hat{X}^{(v)}_{ij} \) as follows:
\begin{equation}\label{updateX}
	\begin{aligned}
		{\hat{X}}^{(v)}_{ij}\!\!\leftarrow~&\bar{M}_{ij}^{(v)}\hat{X}^{(v)}_{ij}\frac{(\omega_v^{\gamma}\bm{W}^{(v)}\bm{\Gamma}^{(v)}+\tau\bm{\hat{X}}^{(v)}\bm{S}^{(v)})_{ij}}{(\omega_v^{\gamma}\bm{W}^{(v)}\bm{W}^{(v)^\mathrm{T}}\bm{\hat{X}}^{(v)}\!\!+\!\tau\bm{\hat{X}}^{(v)}\bm{D}^{(v)})_{ij}}\\
		&+M_{ij}^{(v)}{X}_{ij}^{(v)}
	\end{aligned}
\end{equation}
where $\bar{M}_{ij}^{(v)}=1-M_{ij}^{(v)}$.

\noindent\textbf{Update $\bm{W}^{(v)}$: }
With other variables fixed, the optimization problem w.r.t. $\bm{W}^{(v)}$ is:
\begin{equation}\label{w1}
	\begin{aligned}
		\min_{\bm{W}^{(v)}\geq0} & \|\bm{W}^{(v)\mathrm{T}}\bm{\hat{X}}^{(v)}-\bm{\Gamma}^{(v)}\|_{F}^{2}+\lambda\|\bm{W}^{(v)}\|_{2,1}\\
	\end{aligned}
\end{equation}
where $\bm{\Gamma}^{(v)}=\bm{A}diag(\bm{P}_{v.})\bm{H}^\mathrm{T}$. Following \cite{l21}, we replace $\|\bm{W}^{(v)}\|_{2,1}$ with  $\operatorname{Tr}(\bm{W}^{(v)\mathrm{T}}\bm{\Lambda}^{(v)}\bm{W}^{(v)})$, where $\bm{\Lambda}^{(v)}$ is a diagonal matrix whose $i$-th diagonal entry is defined as $1/(2\|\bm{W}_{i.}^{(v)}\|_2 + \epsilon)$. Then, the update rule for \(\bm{W}^{(v)}\) is given as follows:

\begin{equation}\label{updateW}
	\begin{aligned}
		W^{(v)}_{ij}\leftarrow W^{(v)}_{ij}\frac{(\bm{\hat{X}}^{(v)}\bm{\Gamma}^{(v)})_{ij}}{(\bm{\hat{X}}^{(v)}\bm{\hat{X}}^{(v)\mathrm{T}}\bm{W}^{(v)}+\lambda\bm{\Lambda}^{(v)}\bm{W}^{(v)})_{ij}}
	\end{aligned}
\end{equation}

\noindent\textbf{Update $\bm{A}$: }
After fixing other variables, the objective function w.r.t. $\bm{A}$ is reduced to:
\begin{equation}\label{A1}
	\begin{aligned}
		\min_{\bm{A}\geq0} & \|\bm{\widetilde{\bm{\mathcal{Z}}}}_{(1)}-\bm{A}(\bm{\widetilde{P}}\odot \bm{H})^\mathrm{T}\|_F^2\\
	\end{aligned}
\end{equation}
where $\bm{\widetilde{P}} = \bm{P}_{\bm{\omega},{\gamma}}$, $\bm{\widetilde{\bm{\mathcal{Z}}}} = \bm{\bm{\mathcal{Z}}}_{\bm{\omega},{\gamma}}$, and $\bm{\widetilde{\bm{\mathcal{Z}}}}_{(1)}\in\mathbb{R}^{c\times nV}$ is the mode-1 unfolding of tensor $\bm{\widetilde{\bm{\mathcal{Z}}}}$. By differentiating Eq.~(\ref{A1}) w.r.t. $\bm{A}$, we update $\bm{A}$ as follows:
\begin{equation}\label{updateA}
	\begin{aligned}
		A_{ij}\leftarrow A_{ij}\frac{[\bm{\widetilde{\bm{\mathcal{Z}}}}_{(1)}(\bm{\widetilde{P}}\odot\bm{H})]_{ij}}{[\bm{A}(\bm{\widetilde{P}}\odot\bm{H})^{\mathrm{T}}(\bm{\widetilde{P}}\odot\bm{H})]_{ij}}
	\end{aligned}
\end{equation}

\noindent\textbf{Update $\bm{P}$ and $\bm{\hat{b}}_v$: }
Fixing all other variables, the objective function w.r.t. $\bm{P}$ can be written as:
\begin{equation}\label{P3}
	\begin{aligned}
		&\min_{\bm{P}\geq0} \|\bm{\bm{\mathcal{Z}}}_{(3)}-\bm{P}(\bm{H}\odot\bm{A})^\mathrm{T}\|_{F}^{2}
	\end{aligned}
\end{equation}
where $\bm{{\bm{\mathcal{Z}}}}_{(3)}\in\mathbb{R}^{V\times nc}$ is the mode-3 unfolding of tensor $\bm{{\bm{\mathcal{Z}}}}$. Similar to updating  $\bm{A}$, the update rule for $\bm{P}$ is as follows:
\begin{equation}\label{updateP}
	\begin{aligned}
		P_{ij}\leftarrow P_{ij}\frac{[\bm{\mathcal{Z}}_{(3)}(\bm{H}\odot\bm{A})]_{ij}}{[\bm{P}(\bm{H}\odot\bm{A})^{\mathrm{T}}(\bm{H}\odot\bm{A})]_{ij}}
	\end{aligned}
\end{equation}

The belief mass \( \bm{\hat{b}}_v \) is then updated according to Eq.~(\ref{be2}).

\noindent\textbf{Update $\bm{S}^{(v)}$: }
With other variables fixed, optimizing  $\bm{S}^{(v)}$ reduces to solving the following column-wise subproblem:
\begin{equation}\label{S3}
	\begin{aligned}
		\min_{\bm{S}^{(v)}_{.i}}~&\|\bm{S}^{(v)}_{.i}\!-\!\bm{Q}^{(v)}_{.i}\|_{2}^{2},~
		s.t. \bm{S}^{(v){\mathrm{T}}}_{.i}\bm{1}=1,\bm{S}^{(v)}_{.i}\geq0,{S}^{(v)}_{ii}=0,
	\end{aligned}
\end{equation}
where $\bm{Q}^{(v)}=\frac{\bm{C}^{(v)}+\sum_{k=1,k\neq v}^{V}b_{kv}\bm{R}^{(k)}-\bm{F}/2}{1+\sum_{k=1,k\neq v}^{V}b_{kv}^{2}}$, $\bm{R}^{(k)}=\bm{S}^{(k)}-\sum^V_{t=1,t\neq k,t\neq v}b_{kt}\bm{S}^{(t)}$, $F^{(v)}_{ij}=\|\bm{\hat{X}}^{(v)}_{.i}\!\!-\bm{\hat{X}}^{(v)}_{.j}\|_{2}^{2}+\|\bm{H}_{i.}\!\!-\bm{H}_{j.}\|_{2}^{2}$, and $\bm{C}^{(v)}=\sum_{k=1,k\neq v}^{V}b_{vk}\bm{S}^{(k)}$. This problem can be addressed by following the solution strategy in \cite{GMC}.

\noindent\textbf{Update $\bm{H}$: }
As with the update of  $\bm{A}$, we derive the following equivalent optimization problem w.r.t. $\bm{H}$ by fixing the remaining variables:
\begin{equation}\label{H1}
	\begin{aligned}
		\min_{\bm{H}\geq0} & \|\bm{\widetilde{{\mathcal{Z}}}}_{(2)}-\bm{H}(\bm{\widetilde{P}}\odot\bm{A})^\mathrm{T}\|_F^2+\tau\sum_{v=1}^{V}\operatorname{Tr}(\bm{H}^{\mathrm{T}}\bm{L}^{(v)}\bm{H}),\\
	\end{aligned}
\end{equation}
where $\bm{\widetilde{{\mathcal{Z}}}}_{(2)} \in \mathbb{R}^{n \times Vc}$ denotes the mode-2 unfolding of the tensor $\bm{\widetilde{\bm{\mathcal{Z}}}}$. The update rule for \( \bm{H} \) is then given by:
\begin{equation}\label{updateH}
	\begin{aligned}
		H_{ij}\leftarrow H_{ij}\frac{[\bm{\widetilde{\mathcal{Z}}}_{(2)}(\bm{\widetilde{P}}\odot\bm{A})+\tau\sum_{v=1}^{V}\bm{S}^{(v)}\bm{H}]_{ij}}{[\bm{H}(\bm{\widetilde{P}}\odot\bm{A})^{\mathrm{T}}(\bm{\widetilde{P}}\odot\bm{A})+\tau\sum_{v=1}^{V}\bm{D}^{(v)}\bm{H}]_{ij}}
	\end{aligned}
\end{equation}

\noindent\textbf{Update $\bm{\omega}$: }
By fixing the other variables, we derive the closed-form solution for $\omega_v$ using the procedure in~\cite{TIME-FS}, as follows:
\begin{equation}\label{updateAlpha}
	\begin{aligned}
		\omega_{v}=\frac{(b^{(v)})^{\frac{1}{1-\gamma}}}{\sum_{v=1}^{V}(b^{(v)})^{\frac{1}{1-\gamma}}}
	\end{aligned}
\end{equation}	
where $b^{(v)}\!\!=\!\!\|\!\bm{W}^{(v)\mathrm{T}}{\bm{\hat{X}}}^{(v)}\!-\!\bm{A}diag(\bm{P}_{v.})\bm{H}^{\mathrm{T}}\!\|_{F}^{2}\!+\!\lambda\|\!\bm{W}^{(v)}\!\|_{2,1}$.

The optimization procedure of TRUST-FS is summarized  in Algorithm~\ref{alg:algorithm-1}. In this algorithm, $\bm{\hat{X}}^{(v)}$ is initialized by imputing the missing values with the feature-wise mean of the corresponding view. The matrices $\{\bm{W}^{(v)}\}_{v=1}^{V}$, $\bm{A}$, $\bm{P}$ and $\bm{H}$ are randomly initialized, subject to non-negativity constraints. The similarity matrices $\{\bm{S}^{(v)}\}_{v=1}^{V}$ are initially constructed using a $k$-nearest neighbor graph. The weight vector $\bm{\omega}$ is initialized according to Eq.~(\ref{updateAlpha}).
\begin{algorithm}[!t]
	\caption{Iterative Algorithm of TRUST-FS}
	\label{alg:algorithm-1}
	\textbf{Input}: Incomplete multi-view data $\bm{\mathcal{X}}=\{\bm{X}^{(v)}\}_{v=1}^V$, and parameters $\gamma$, $\lambda$, and $\tau$.
	\begin{algorithmic}[1]
		\STATE Initialize $\{\bm{\hat{X}}^{(v)}, \bm{W}^{(v)}, \bm{S}^{(v)}\}_{v=1}^{V}$, $\bm{A}$, $\bm{P}$, $\bm{H}$, $\bm{\hat{b}}$ and $\bm{\omega}$.
		\WHILE{not convergent}
		\STATE Update $\{\bm{\hat{X}}^{(v)}\}_{v=1}^{V}$ via Eq.~(\ref{updateX});
		\STATE Update $\{\bm{W}^{(v)}\}_{v=1}^{V}$ via Eq.~(\ref{updateW});
		\STATE Update $\bm{A}$ via Eq.~(\ref{updateA});
		\STATE Update $\bm{P}$ via Eq.~(\ref{updateP}), and update $\bm{\hat{b}}$ by Eq.~(\ref{be2});
		\STATE Update $\{\bm{S}^{(v)}\}_{v=1}^{V}$ by solving Eq.~(\ref{S3});
		\STATE Update $\bm{H}$ via Eq.~(\ref{updateH});		 
		\STATE Update $\bm{\omega}$ via Eq.~(\ref{updateAlpha});
		\ENDWHILE
	\end{algorithmic}
	\textbf{Output}: {Sorting the $\ell_{2}$-norm of the rows of $\{\mathbf{W}^{(v)} \}_{v=1}^{V}$ in descending order and selecting the top $l$ features.}
\end{algorithm}
\section{Convergence and Complexity Analysis}
\subsection{Convergence Analysis}

In this section, we present a theoretical proof of the convergence of Algorithm~\ref{alg:algorithm-1}. Because the objective function is jointly non-convex with respect to all variables, Eq.~(\ref{final_obj}) is decomposed into seven subproblems. We then demonstrate that the value of the objective function decreases monotonically with each iteration. Specifically, we denote the objective function in Eq.~(\ref{final_obj}) as $\mathcal{J}(\hat{\bm{X}}^{(v)}, \bm{W}^{(v)}, \bm{S}^{(v)}, \bm{A}, \bm{P}, \bm{\hat{b}}_v, \bm{H}, \bm{\omega})$, and represent the solution at iteration $t$ as $\bm{W}^{(v)}_t$, $\hat{\bm{X}}^{(v)}_t$, $\bm{S}^{(v)}_t$, $\bm{A}_t$, $\bm{P}_t$, $\bm{\hat{b}}_{v,t}$, $\bm{H}_t$ and $\bm{\omega}_t$.

\begin{theorem}\label{Theorem1}
	The update rules in Algorithm~1 guarantee that the objective function $\mathcal{J}(\hat{\bm{X}}^{(v)}, \bm{W}^{(v)}, \bm{S}^{(v)}, \bm{A}, \bm{P}, \bm{\hat{b}}_v, \bm{H}, \bm{\omega})$ decreases monotonically with each iteration until convergence.
\end{theorem}

\begin{proof}The proof of Theorem~\ref{Theorem1} is divided into three parts, as described below.

\noindent\textbf{1) $\mathcal{J}(\hat{\bm{X}}^{(v)}, \bm{W}^{(v)}, \bm{S}^{(v)}, \bm{A}, \bm{P}, \bm{\hat{b}}_v, \bm{H}, \bm{\omega})$ is bounded below:} 
We begin by demonstrating that when $\bm{S}^{(v)} \geq 0$, the term $\operatorname{Tr}(\bm{H}^\mathrm{T}\bm{L}^{(v)}\bm{H})$ is bounded below. 
By rewriting the trace term, we obtain
\[
\operatorname{Tr}(\bm{H}^\mathrm{T}\bm{L}^{(v)}\bm{H}) 
= \frac{1}{2}\sum_{i,j=1}^{n}\|\bm{H}_{i.}-\bm{H}_{j.}\|_2^2 S_{ij}^{(v)}.
\]
Since $S_{ij}^{(v)} \geq 0$, it follows that $\operatorname{Tr}(\bm{H}^\mathrm{T}\bm{L}^{(v)}\bm{H}) \geq 0$. 
Moreover, the remaining terms in the objective function involve either the $\ell_{2,1}$-norm or the Frobenius norm, both of which are nonnegative. 
Consequently, the overall objective function
$\mathcal{J}(\hat{\bm{X}}^{(v)}, \bm{W}^{(v)}, \bm{S}^{(v)}, \bm{A}, \bm{P}, \bm{\hat{b}}_v, \bm{H}, \bm{\omega})$ 
is bounded below by zero.

\noindent\textbf{2) Updating $\bm{W}^{(v)}$, $\bm{\hat{X}}^{(v)}$, $\bm{A}$, $\bm{P}$, $\bm{H}$, $\bm{\hat{b}}_v$:} 
To demonstrate the convergence of Algorithm~1, we first show that the objective function in Eq.~(\ref{w1}) decreases monotonically with respect to updates of $\bm{W}^{(v)}$, while keeping all other variables fixed. Following the approach in \cite{l21}, the subproblem in (\ref{w1}) can be reformulated as

\begin{equation}~\label{neq19}
	\begin{aligned}
		\min_{\bm{W}^{(v)}\geq 0} 
		\|\bm{W}^{(v)\mathrm{T}}\bm{\hat{X}}^{(v)}-\bm{\Gamma}^{(v)}\|_{F}^{2}
		+\lambda\,\operatorname{Tr}(\bm{W}^{(v)\mathrm{T}}\bm{\Lambda}^{(v)}\bm{W}^{(v)}).
	\end{aligned}
\end{equation}
Let the objective function in (\ref{neq19}) be denoted by $\mathcal{J}_{\bm{W}}(\bm{W}^{(v)})$. By using the identity $\|\bm{A}\|_F^2 = \operatorname{Tr}(\bm{A}^{\mathrm{T}}\bm{A})$, we can rewrite $\mathcal{J}_{\bm{W}}(\bm{W}^{(v)})$ as
\begin{equation}
	\begin{aligned}
		&\mathcal{J}_{\bm{W}}(\!\bm{W}^{(v)}\!)=\operatorname{Tr}(\bm{\hat{X}}^{(v)\mathrm{T}}\bm{W}^{(v)}\bm{W}^{(v)\mathrm{T}}\bm{\hat{X}}^{(v)})+\operatorname{Tr}(\bm{\Gamma}^{(v)\mathrm{T}}\bm{\Gamma}^{(v)})\\
		&-2\operatorname{Tr}(\bm{\Gamma}^{(v)\mathrm{T}}\bm{W}^{(v)\mathrm{T}}\bm{\hat{X}}^{(v)})+\lambda\operatorname{Tr}(\bm{W}^{(v)\mathrm{T}}\bm{\Lambda}^{(v)}\bm{W}^{(v)})
	\end{aligned}
\end{equation}

To facilitate the following proof, we adopt the auxiliary function approach. Specifically, we present the following definition and lemmas from \cite{convergence}, which are essential for constructing an appropriate auxiliary function for $\mathcal{J}_{\bm{W}}$.

\begin{definition}[]
	\cite{convergence} $\mathcal{G}(u,u^{\prime})$  is called an auxiliary function for  $\mathcal{J}(u)$ if
	$$
	\mathcal{J}(u)\leq\mathcal{G}(u,u^{\prime}), \quad \mathcal{G}(u,u)=\mathcal{J}(u).
	$$
\end{definition}

\begin{lemma}[]\label{Lemma1}
	\cite{convergence} If $\mathcal{G}$ is an auxiliary function of $\mathcal{J}$, then $\mathcal{J}$  is non-increasing under the update rule
	$$
	u_{t+1}=\arg\min_{u}\mathcal{G}(u,u_t).
	$$
\end{lemma}

\begin{lemma}[]\label{Lemma2}
	\cite{convergence} For any nonnegative matrices $\bm{A}\in\mathbb{R}^{n\times n}$, $\bm{B}\in\mathbb{R}^{k\times k}$, $\bm{S}\in\mathbb{R}^{n\times k}$, $ \bm{S^{\prime}}\in\mathbb{R}^{n\times k}$, if $\bm{A}$ and $\bm{B}$ are symmetric, then
	$$
	\operatorname{Tr}(\bm{S}^{\mathrm{T}}\bm{A}\bm{S}\bm{B})\leq
	\sum_{i=1}^n\sum_{j=1}^k
	\frac{(\bm{A}\bm{S}^{\prime}\bm{B})_{ij}S^2_{ij}}{S_{ij}^{\prime}}.
	$$
\end{lemma}
Based on Lemma~\ref{Lemma2} and the cyclic property of the trace operator, we have
\begin{equation}\label{c1}
	\begin{aligned}
		&\operatorname{Tr}(\bm{\hat{X}}^{(v)\mathrm{T}}\!\bm{W}^{(v)}\!\bm{W}^{(v)\mathrm{T}}\!\bm{\hat{X}}^{(v)}\!)\!
		=\!\operatorname{Tr}(\bm{W}^{(v)\mathrm{T}}\!(\bm{\hat{X}}^{(v)}\!\bm{\hat{X}}^{(v)\mathrm{T}})\!\bm{W}^{(v)}\!\bm{I}_{c}\!)\\
		&\leq\sum_{i=1}^{d_v}\sum_{j=1}^{c}(\bm{\hat{X}}^{(v)}\bm{\hat{X}}^{(v)\mathrm{T}}\bm{W}^{(v)\prime})_{ij}\frac{W^{(v)2}_{ij}}{W^{(v)\prime}_{ij}}
	\end{aligned}
\end{equation}	

Similarly,
\begin{equation}\label{c2}
	\operatorname{Tr}(\bm{W}^{(v)\mathrm{T}}\bm{\Lambda}^{(v)}\bm{W}^{(v)})
	\leq \sum_{i=1}^{d_v}\sum_{j=1}^c
	\frac{(\bm{\Lambda}^{(v)}\bm{W}^{(v)\prime})_{ij}\,W^{(v)2}_{ij}}{W^{(v)\prime}_{ij}}.
\end{equation}

In addition, the term $\operatorname{Tr}(\bm{\Gamma}^{(v)\mathrm{T}}\bm{W}^{(v)\mathrm{T}}\bm{\hat{X}}^{(v)})$ can be expanded element-wise as follows:
\begin{equation}
	\begin{aligned}\label{c3}
		&\operatorname{Tr}(\bm{\Gamma}^{(v)\mathrm{T}}\bm{W}^{(v)\mathrm{T}}\bm{\hat{X}}^{(v)})
		=\sum_{i=1}^{d_v}\sum_{j=1}^c(\bm{\hat{X}}^{(v)}\bm{\Gamma}^{(v)\mathrm{T}})_{ij}W^{(v)}_{ij}
	\end{aligned}
\end{equation}	

Based on Eqs. (\ref{c1})–(\ref{c3}), we construct the auxiliary function $\mathcal{G}(\bm{W}^{(v)},\bm{W}^{(v)\prime})$ as follows:
\begin{equation}
	\begin{aligned}\label{c4}
		&\mathcal{G}(\bm{W}^{(v)},\bm{W}^{(v)^\prime})=\sum_{i=1}^{d_v}\sum_{j=1}^{c}(\bm{\hat{X}}^{(v)}\bm{\hat{X}}^{(v)\mathrm{T}}\bm{W}^{(v)\prime})_{ij}\frac{W^{(v)2}_{ij}}{W^{(v)\prime}_{ij}}\\
		&+\operatorname{Tr}(\bm{\Gamma}^{(v)\mathrm{T}}\bm{\Gamma}^{(v)})+\sum_{i=1}^{d_v}\sum_{j=1}^c\lambda(\bm{\Lambda}^{(v)}\bm{W}^{(v)\prime}_{ij})_{ij}\frac{W^{(v)2}_{ij}}{W^{(v)\prime}_{ij}}\\
		&-2\sum_{i=1}^{d_v}\sum_{j=1}^c(\bm{\hat{X}}^{(v)}\bm{\Gamma}^{(v)\mathrm{T}})_{ij}W^{(v)}_{ij}
	\end{aligned}
\end{equation}	

It is straightforward to verify that $\mathcal{G}(\bm{W}^{(v)}, \bm{W}^{(v)\prime}) \geq \mathcal{J}_{\bm{W}}(\bm{W}^{(v)})$ and $\mathcal{G}(\bm{W}^{(v)}, \bm{W}^{(v)\prime}) = \mathcal{J}_{\bm{W}}(\bm{W}^{(v)})$. Therefore, by Definition~1, $\mathcal{G}(\bm{W}^{(v)}, \bm{W}^{(v)\prime})$ qualifies as an auxiliary function for  $\mathcal{J}_{\bm{W}}(\bm{W}^{(v)})$.

Taking the first derivative of $\mathcal{G}(\bm{W}^{(v)}, \bm{W}^{(v)\prime})$ w.r.t. $W^{(v)}_{ij}$ yields
\begin{equation}
	\begin{aligned}\label{c5}
		&\frac{\partial\mathcal{G}(\bm{W}^{(v)},\bm{W}^{(v)^\prime})}{\partial W^{(v)}_{ij}}=2(\bm{\hat{X}}^{(v)}\bm{\hat{X}}^{(v)\mathrm{T}}\bm{W}^{(v)\prime})_{ij}\frac{W^{(v)}_{ij}}{W^{(v)\prime}_{ij}}\\
		&+2\lambda(\bm{\Lambda}^{(v)}\bm{W}^{(v)\prime})_{ij}\frac{W^{(v)}_{ij}}{W^{(v)\prime}_{ij}}-2(\bm{\hat{X}}^{(v)}\bm{\Gamma}^{(v)\mathrm{T}})_{ij}
	\end{aligned}
\end{equation}	

The second derivative w.r.t. $W^{(v)}_{ij}$ and $W^{(v)}_{pq}$ is given by
\begin{equation}
	\begin{aligned}\label{c6}
		&\frac{\partial^2\mathcal{G}(\bm{W}^{(v)},\bm{W}^{(v)^\prime})}{\partial W^{(v)}_{ij}\partial W^{(v)}_{pq}}=2\delta_{ip}\delta_{jq}[(\bm{\hat{X}}^{(v)}\bm{\hat{X}}^{(v)\mathrm{T}}\bm{W}^{(v)\prime})_{ij}\frac{1}{W^{(v)\prime}_{ij}}\\
		&+2\lambda(\bm{\Lambda}^{(v)}\bm{W}^{(v)\prime})_{ij}\frac{1}{W^{(v)\prime}_{ij}}]\geq0
	\end{aligned}
\end{equation}	
where $\delta_{ip}=1$ if $i=p$, and $\delta_{ip}=0$ otherwise. The resulting Hessian matrix is diagonal with positive entries, confirming  that $\mathcal{G}(\bm{W}^{(v)}, \bm{W}^{(v)\prime})$ is convex w.r.t. $\bm{W}^{(v)}$. 

Consequently, the global minimum can be found by setting
$\frac{\partial \mathcal{G}(\bm{W}^{(v)}, \bm{W}^{(v)\prime})}{\partial W^{(v)}_{ij}}=0$.
Solving this equation yields the optimal $\bm{W}^{(v)}$, which corresponds to the update rule given in Eq.~(\ref{updateW}). Hence, the update rule for $\bm{W}^{(v)}$ in Algorithm~1  coincides with the minimizer of the auxiliary function:
\[
\bm{W}_{t+1}^{(v)}=\arg\min_{\bm{W}^{(v)}}\mathcal{G}(\bm{W}^{(v)},\bm{W}^{(v)}_t).
\]

Since $\mathcal{G}(\bm{W}^{(v)}, \bm{W}^{(v)}_t)$ serves as an auxiliary function for $\mathcal{J}_{\bm{W}}(\bm{W}^{(v)})$, Lemma~\ref{Lemma1} guarantees that
\begin{equation}\label{c7}
	\mathcal{J}_{\bm{W}}(\bm{W}^{(v)}_{t+1})\!\!
	\leq\!
	\mathcal{G}\!(\bm{W}^{(v)}_{t+1},\!\bm{W}^{(v)}_t\!)\!\!
	\leq\!
	\mathcal{G}\!(\bm{W}^{(v)}_{t}\!,\!\bm{W}^{(v)}_t\!)\!\!
	=\!\mathcal{J}_{\bm{W}}(\bm{W}^{(v)}_{t}\!),
\end{equation}
thereby ensuring that the objective function in Eq.~(\ref{w1}) decreases monotonically with each iteration of Algorithm~1.

The monotonic convergence of $\bm{\hat{X}}^{(v)}$, $\bm{A}$, $\bm{P}$, and $\bm{H}$ can be established analogously. In addition, as $\bm{\hat{b}}_v$  is explicitly computed from the updated $\bm{P}$ at each iteration, it does not constitute an independent optimization variable. Consequently, the convergence of $\bm{P}$  directly implies the convergence of $\bm{\hat{b}}_v$.

\noindent\textbf{3) Updating $\bm{S}^{(v)}$, $\bm{\omega}$:} Since both $\bm{S}^{(v)}$ and $\bm{\omega}$ are updated using the closed-form optimal solutions of their respective subproblems, we have 

\begin{equation}\label{c8}
	\begin{aligned}
		\mathcal{J}(\hat{\bm{X}}^{(v)}_{t+1}, \bm{W}^{(v)}_{t+1}, \bm{S}^{(v)}_t, \bm{A}_{t+1}, \bm{P}_{t+1}, \bm{\hat{b}}_{v,t+1}, \bm{H}_{t}, \bm{\omega}_t)\leq\\
		\mathcal{J}(\hat{\bm{X}}^{(v)}_{t+1}, \bm{W}^{(v)}_{t+1}, \bm{S}^{(v)}_{t+1}, \bm{A}_{t+1}, \bm{P}_{t+1}, \bm{\hat{b}}_{v,t+1}, \bm{H}_{t}, \bm{\omega}_{t})
	\end{aligned}
\end{equation}	
and
\begin{equation}\label{c9}
	\begin{aligned}
		\mathcal{J}(\hat{\bm{X}}^{(v)}_{t+1}, \bm{W}^{(v)}_{t+1}, \bm{S}^{(v)}_{t+1}, \bm{A}_{t+1}, \bm{P}_{t+1}, \bm{\hat{b}}_{v,t+1}, \bm{H}_{t+1}, \bm{\omega}_t)\leq\\
		\mathcal{J}(\hat{\bm{X}}^{(v)}_{t+1}, \bm{W}^{(v)}_{t+1}, \bm{S}^{(v)}_{t+1}, \bm{A}_{t+1}, \bm{P}_{t+1}, \bm{\hat{b}}_{v,t+1}, \bm{H}_{t+1}, \bm{\omega}_{t+1})
	\end{aligned}
\end{equation}	

Therefore, Algorithm~1 guarantees that the objective value decreases monotonically and converges, as the iterative sequence is non-increasing and bounded below.

\subsection{Complexity Analysis}
In Algorithm~\ref{alg:algorithm-1}, updating $\bm{\hat{X}}^{(v)}$ mainly involves matrix multiplication, resulting in a time complexity of $\mathcal{O}(d_vnc + d_vn^2)$. Similarly, updating $\bm{W}^{(v)}$ has a time complexity of $\mathcal{O}(d_vnc)$. The time complexity for updating each of $\bm{A}$, $\bm{H}$, and $\bm{P}$ is $\mathcal{O}(crnV)$. For updating $\bm{S}^{(v)}$, it takes $\mathcal{O}(d_vn^2)$ complexity. The updates of $\bm{\omega}$ and $\bm{\hat{b}}$ involve only element-wise operations and are therefore negligible. Consequently, the total time complexity of Algorithm~\ref{alg:algorithm-1} is $\mathcal{O}(dnc + crnV + dn^2)$, where $d = \sum_{v=1}^{V} d_v$. 
\end{proof}

\section{Experiment}

In this section, we conduct extensive experiments to evaluate the effectiveness of TRUST-FS by comparing it with several state-of-the-art baseline methods.

\subsection{Experimental Schemes}
\subsubsection{datasets}
We evaluate the effectiveness of our method on eight widely used multi-view datasets. The description of these datasets are presented as follows:

\textit{Football} \cite{rugby} consists of 248 English Premier League players and clubs active on Twitter, organized into 20 distinct ground-truth communities.

\textit{ORL} \cite{missing-variable2} is a classic face recognition dataset comprising 400 grayscale images of 40 subjects, captured with different expressions and lighting conditions across three views.

\textit{Rugby} \cite{rugby} includes 854 Rugby Union players, clubs, and organizations on Twitter, forming 15 national communities.

\textit{100leaves} \cite{100leaves}, collected from the UCI repository, contains 1,600 leaf images from 100 plant species.

\textit{BDGP} \cite{BDGP} comprises 2,500 images of Drosophila embryos, grouped into five distinct categories.

\textit{Scene} \cite{Scene} contains  4,485 natural images from 15 scene categories, with each category represented by three views.

\textit{MNIST} \cite{mnist} is a handwritten digit dataset consisting of 10,000 grayscale images, evenly divided among ten categories (0–9).

\textit{Aloi} \cite{ALOI} contains images of small objects captured under various viewing angles and illumination conditions, each represented by four types of visual features.

\begin{table}[!t]
	\centering
	\setlength{\tabcolsep}{1.3mm}
	\small
	\caption{A detail description of multi-view datasets}\label{table1}
    \vspace*{-5pt}
	\begin{tabular}{@{\extracolsep{\fill}}lcccc}
		\toprule
		Datasets &   Views & Samples & Features & Classes\\
		\midrule
		\multirow{2}{*}{Football} & \multirow{2}{*}{9} & \multirow{2}{*}{248} & 11806/7814/3601/248/ & \multirow{2}{*}{20} \\
		& & & 248/248/248/248/248 &\\
		ORL & 3 & 400 & 4096/3304/6750 & 40 \\
		\multirow{2}{*}{Rugby} & \multirow{2}{*}{9} & \multirow{2}{*}{854} & 35352/5900/2840/854/ & \multirow{2}{*}{15} \\
		& & & 854/854/854/854/854 &\\
		100leaves & 3 &1600 & 64/64/64& 100 \\
		BDGP & 3 & 2500 & 100/500/250 & 5 \\
		Scene & 3 &4485 &  20/59/40 & 15 \\
		MNIST &3 &10000&30/9/30&10\\
		Aloi & 4 & 11025 & 77/13/64/64&100\\
		\toprule
	\end{tabular}
	
\end{table}

The detail statistic descriptions of these datasets are provided in Table~\ref{table1}. Additionally, we generated incomplete multi-view datasets with missing variables by randomly removing a certain proportion of entries, following\cite{simulate-missing-variable}. The missing ratio is set in increments of 20\%, ranging from 10\% to 90\% in the experiment.

\subsubsection{Evaluation Metrics}

Following previous studies on MUFS~\cite{C2IMUFS,Zhang2024Scalable}, the quality of selected features is evaluated using Clustering Accuracy (ACC) and Normalized Mutual Information (NMI), which are defined as follows.
\begin{equation}\label{acc}
	\begin{aligned}
		ACC = \frac{1}{n}\sum_{i=1}^{n}\delta(y_i,\text{map}(\hat{y}_{i}))
	\end{aligned}
\end{equation}
where $y_i$ and $\hat{y}_i$ represent the ground-truth label and clustering assignment of the $i$-th instance, respectively; $n$ denotes the total number of samples; $\delta(x,y)$ is an indicator function that equals $1$ if $x=y$ and $0$ otherwise; and $\text{map}(\cdot)$  is a permutation mapping function that aligns each cluster index with the optimal ground-truth label using the Kuhn–Munkres algorithm~\cite{bestmap}.

Let $\mathcal{Y} = \{\mathcal{Y}_i\}_{i=1}^c$ and $\mathcal{Y}^{\prime} = \{\mathcal{Y}_j^{\prime}\}_{j=1}^{c^{\prime}}$ denote the ground-truth partition and the resulting clustering, respectively. The Normalized Mutual Information (NMI) is then defined as:
\begin{equation}\label{NMI}
	\begin{aligned}
		NMI(\mathcal{Y},\mathcal{Y}^{\prime}) = \frac{\sum_{i=1}^c\sum_{j=1}^{c^{\prime}}|\mathcal{Y}_i\cap\mathcal{Y}_j^{{\prime}}|log\frac{n|\mathcal{Y}_i\cap\mathcal{Y}_j^{{\prime}}|}{|\mathcal{Y}_i||\mathcal{Y}_j^{{\prime}}|}}{max(\sum_{i=1}^c|\mathcal{Y}_i|\frac{|\mathcal{Y}_i|}{n},\sum_{j=1}^{c^{\prime}}|\mathcal{Y}_j^{{\prime}}|log\frac{|\mathcal{Y}_j^{{\prime}}|}{n})}
	\end{aligned}
\end{equation}
where $|\cdot|$ the cardinality of a set.

For both metrics, higher values indicate better clustering performance.

\begin{table*}[t]\huge 
	\tabcolsep 0pt
	\caption{Means and standard deviation (\%) of ACC for different methods on eight datasets with a missing ratio of 50\% while selecting 30\% of all features.}\small\label{table2}
	\vspace*{-15pt}
	\begin{flushleft}
		\def\temptablewidth{\textwidth}
		{\rule{\temptablewidth}{1pt}}
		\begin{tabular*}{\temptablewidth}{@{\extracolsep{\fill}}lcccccccc}
			\diagbox{Methods}{Datasets} & Football  &  ORL & Rugby & 100leaves & BDGP  & Scene  & MNIST & Aloi \\
			\hline
            OURS & $\mathbf{37.10\pm 6.83}$ & $\mathbf{55.64\pm 3.66}$ & $\mathbf{50.53\pm 8.39}$ & $\mathbf{63.06\pm 2.56}$ & $\mathbf{47.19\pm 3.80}$ & $\mathbf{38.81\pm 1.57}$ & $\mathbf{68.83\pm 6.16}$ & $\mathbf{57.42\pm 2.33}$ \\
            AllFea & $15.04\pm 2.08\,\bullet$ & $43.46\pm 3.96\,\bullet$ & $37.38\pm 2.73\,\bullet$ & $\underline{49.26\pm 2.06}\,\bullet$ & $42.10\pm 3.56\,\bullet$ & $\underline{35.32\pm 1.95}\,\bullet$ & $\underline{54.17\pm 1.52}\,\bullet$ & $\underline{39.10\pm 1.73}\,\bullet$ \\
            C$^2$IMUFS & $20.58\pm 5.46\,\bullet$ & $41.73\pm 2.68\,\bullet$ & $40.80\pm 6.14\,\bullet$ & $26.87\pm 0.80\,\bullet$ & $43.16\pm 5.39\,\bullet$ & $30.60\pm 0.88\,\bullet$ & $29.08\pm 2.34\,\bullet$ & $31.14\pm 0.95\,\bullet$ \\
            CE-UMFS & $21.03\pm 5.08\,\bullet$ & $47.84\pm 4.52\,\bullet$ & $\underline{44.20\pm 5.77}\,\bullet$ & $31.90\pm 1.03\,\bullet$ & $\underline{43.19\pm 4.17}\,\bullet$ & $28.56\pm 1.19\,\bullet$ & $51.23\pm 3.41\,\bullet$ & $32.32\pm 0.97\,\bullet$ \\
            CFSMO & $22.14\pm 7.13\,\bullet$ & $47.06\pm 3.46\,\bullet$ & $41.35\pm 6.33\,\bullet$ & $31.05\pm 1.00\,\bullet$ & $40.59\pm 3.32\,\bullet$ & $24.58\pm 0.97\,\bullet$ & $46.18\pm 4.21\,\bullet$ & $25.08\pm 1.12\,\bullet$ \\
            UNIFIER & $23.12\pm 6.01\,\bullet$ & $\underline{48.44\pm 2.63}\,\bullet$ & $40.96\pm 5.38\,\bullet$ & $31.96\pm 1.15\,\bullet$ & $42.87\pm 3.71\,\bullet$ & $32.09\pm 1.25\,\bullet$ & $48.61\pm 2.76\,\bullet$ & $33.38\pm 1.15\,\bullet$ \\
            WLTL & $21.69\pm 5.09\,\bullet$ & $48.05\pm 3.03\,\bullet$ & $42.27\pm 5.40\,\bullet$ & $31.01\pm 1.18\,\bullet$ & $40.24\pm 2.63\,\bullet$ & $30.11\pm 1.03\,\bullet$ & $45.36\pm 2.62\,\bullet$ & $31.93\pm 0.84\,\bullet$ \\
            UKMFS & $\underline{23.15\pm 7.57}\,\bullet$ & $45.81\pm 3.54\,\bullet$ & $42.27\pm 7.88\,\bullet$ & $30.94\pm 1.35\,\bullet$ & $41.61\pm 5.47\,\bullet$ & $30.10\pm 1.21\,\bullet$ & $42.45\pm 2.96\,\bullet$ & $29.22\pm 0.94\,\bullet$ \\
            GCDUFS & $18.00\pm 4.18\,\bullet$ & $45.82\pm 4.07\,\bullet$ & $37.42\pm 4.30\,\bullet$ & $30.92\pm 0.91\,\bullet$ & $37.83\pm 2.92\,\bullet$ & $25.92\pm 1.12\,\bullet$ & $42.77\pm 2.74\,\bullet$ & $26.51\pm 0.69\,\bullet$ \\
            TIME-FS & $19.48\pm 5.22\,\bullet$ & $40.33\pm 1.88\,\bullet$ & $39.53\pm 4.91\,\bullet$ & $26.82\pm 0.96\,\bullet$ & $39.52\pm 3.41\,\bullet$ & $29.42\pm 1.33\,\bullet$ & $33.07\pm 1.32\,\bullet$ & $30.66\pm 0.76\,\bullet$ \\
            TERUIMUFS & $18.71\pm 5.92\,\bullet$ & $40.60\pm 4.38\,\bullet$ & $38.01\pm 4.19\,\bullet$ & $26.15\pm 1.07\,\bullet$ & $40.13\pm 2.46\,\bullet$ & $28.64\pm 1.47\,\bullet$ & $27.06\pm 1.96\,\bullet$ & $31.00\pm 0.81\,\bullet$ \\
            GAWFS & $14.01\pm 1.23\,\bullet$ & $36.81\pm 2.79\,\bullet$ & $37.86\pm 3.79\,\bullet$ & $29.48\pm 0.84\,\bullet$ & $33.89\pm 3.66\,\bullet$ & $25.12\pm 1.40\,\bullet$ & $43.94\pm 3.77\,\bullet$ & $15.41\pm 0.35\,\bullet$ \\
			\bottomrule
		\end{tabular*}
	\end{flushleft}
\end{table*}
\begin{table*}[t]\huge
	\tabcolsep 0pt
	\caption{Means and standard deviation (\%) of NMI for different methods on eight datasets with a missing ratio of 50\% while selecting 30\% of all features.}\small\label{Table3}
	\vspace*{-15pt}
	\begin{flushleft}
		\def\temptablewidth{\textwidth}
		{\rule{\temptablewidth}{1pt}}
		\begin{tabular*}{\temptablewidth}{@{\extracolsep{\fill}}lcccccccc}
			\diagbox{Methods}{Datasets} & Football  &  ORL & Rugby & 100leaves  & BDGP  & Scene & MNIST & Aloi   \\
			\hline
			OURS & $\mathbf{39.62\pm 7.72}$ & $\mathbf{76.41\pm 2.28}$ & $\mathbf{40.79\pm 8.30}$ & $\mathbf{83.64\pm 0.96}$ & $\mathbf{26.08\pm 3.61}$ & $\mathbf{39.62\pm 0.37}$ & $\mathbf{56.40\pm 1.83}$ & $\mathbf{75.64\pm 0.82}$ \\
            AllFea & $9.81\pm 2.30\,\bullet$ & $61.48\pm 3.82\,\bullet$ & $4.87\pm 4.95\,\bullet$ & $\underline{73.44\pm 1.23}\,\bullet$ & $19.56\pm 5.18\,\bullet$ & $\underline{36.29\pm 0.93}\,\bullet$ & $\underline{47.40\pm 0.93}\,\bullet$ & $\underline{61.12\pm 0.91}\,\bullet$ \\
            C$^2$IMUFS & $17.75\pm 7.14\,\bullet$ & $61.40\pm 1.94\,\bullet$ & $10.76\pm 8.17\,\bullet$ & $56.70\pm 0.61\,\bullet$ & $\underline{21.43\pm 6.48}\,\bullet$ & $31.82\pm 0.65\,\bullet$ & $18.58\pm 1.16\,\bullet$ & $54.43\pm 0.53\,\bullet$ \\
            CE-UMFS & $19.14\pm 7.17\,\bullet$ & $65.14\pm 3.38\,\bullet$ & $15.61\pm 9.18\,\bullet$ & $59.46\pm 0.78\,\bullet$ & $18.56\pm 4.62\,\bullet$ & $27.29\pm 0.55\,\bullet$ & $39.25\pm 1.98\,\bullet$ & $55.26\pm 0.50\,\bullet$ \\
            CFSMO & $20.44\pm 8.74\,\bullet$ & $65.09\pm 2.47\,\bullet$ & $10.70\pm 7.54\,\bullet$ & $58.59\pm 0.65\,\bullet$ & $17.84\pm 4.35\,\bullet$ & $26.15\pm 0.63\,\bullet$ & $36.72\pm 1.81\,\bullet$ & $48.26\pm 0.59\,\bullet$ \\
            UNIFIER & $\underline{21.40\pm 7.78}\,\bullet$ & $\underline{66.83\pm 1.43}\,\bullet$ & $9.33\pm 7.48\,\bullet$ & $59.67\pm 0.86\,\bullet$ & $19.52\pm 3.79\,\bullet$ & $30.24\pm 0.65\,\bullet$ & $40.78\pm 2.28\,\bullet$ & $56.69\pm 0.56\,\bullet$ \\
            WLTL & $20.21\pm 7.48\,\bullet$ & $66.56\pm 2.40\,\bullet$ & $12.50\pm 8.16\,\bullet$ & $59.47\pm 0.60\,\bullet$ & $16.85\pm 3.29\,\bullet$ & $29.48\pm 0.69\,\bullet$ & $38.13\pm 1.64\,\bullet$ & $55.36\pm 0.43\,\bullet$ \\
            UKMFS & $21.17\pm 9.61\,\bullet$ & $64.26\pm 3.01\,\bullet$ & $\underline{18.27\pm 11.90}\,\bullet$ & $58.61\pm 0.67\,\bullet$ & $19.85\pm 6.75\,\bullet$ & $31.74\pm 0.68\,\bullet$ & $34.75\pm 1.47\,\bullet$ & $52.80\pm 0.49\,\bullet$ \\
            GCDUFS & $14.97\pm 5.08\,\bullet$ & $65.30\pm 2.97\,\bullet$ & $4.72\pm 4.49\,\bullet$ & $58.75\pm 0.70\,\bullet$ & $12.59\pm 2.63\,\bullet$ & $26.23\pm 0.86\,\bullet$ & $33.08\pm 0.61\,\bullet$ & $49.84\pm 0.37\,\bullet$ \\
            TIME-FS & $16.20\pm 7.25\,\bullet$ & $60.08\pm 2.31\,\bullet$ & $8.39\pm 4.87\,\bullet$ & $56.78\pm 0.65\,\bullet$ & $15.77\pm 4.48\,\bullet$ & $29.02\pm 0.83\,\bullet$ & $23.16\pm 0.79\,\bullet$ & $54.36\pm 0.52\,\bullet$ \\
            TERUIMUFS & $15.49\pm 8.61\,\bullet$ & $60.61\pm 3.20\,\bullet$ & $5.59\pm 4.33\,\bullet$ & $56.26\pm 0.81\,\bullet$ & $16.99\pm 4.21\,\bullet$ & $29.08\pm 0.95\,\bullet$ & $17.17\pm 1.03\,\bullet$ & $54.41\pm 0.40\,\bullet$ \\
            GAWFS & $9.07\pm 1.07\,\bullet$ & $54.10\pm 3.32\,\bullet$ & $4.41\pm 4.30\,\bullet$ & $56.63\pm 0.75\,\bullet$ & $9.13\pm 3.03\,\bullet$ & $21.13\pm 0.83\,\bullet$ & $37.46\pm 2.37\,\bullet$ & $31.50\pm 0.57\,\bullet$ \\
			\bottomrule
		\end{tabular*}
	\end{flushleft}
\end{table*}

\begin{figure*}[t]
	\centering
	\includegraphics[width=\textwidth]{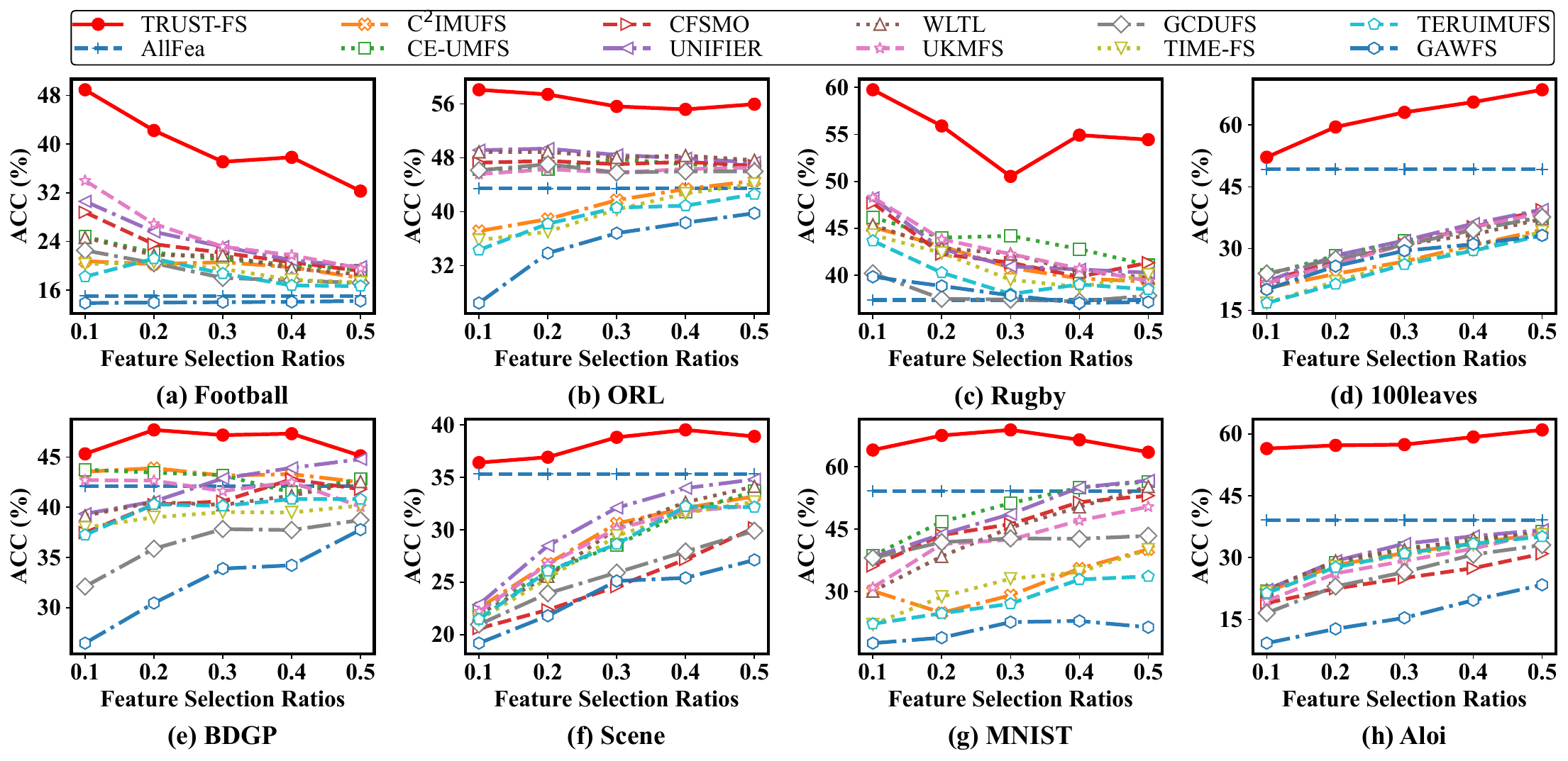}
	\caption{ACC of different methods on eight datasets under different feature selection ratios.}\label{ACC_feature_varying}
\end{figure*}

\begin{figure*}[!htbp]
	\centering
	\includegraphics[width=\textwidth]{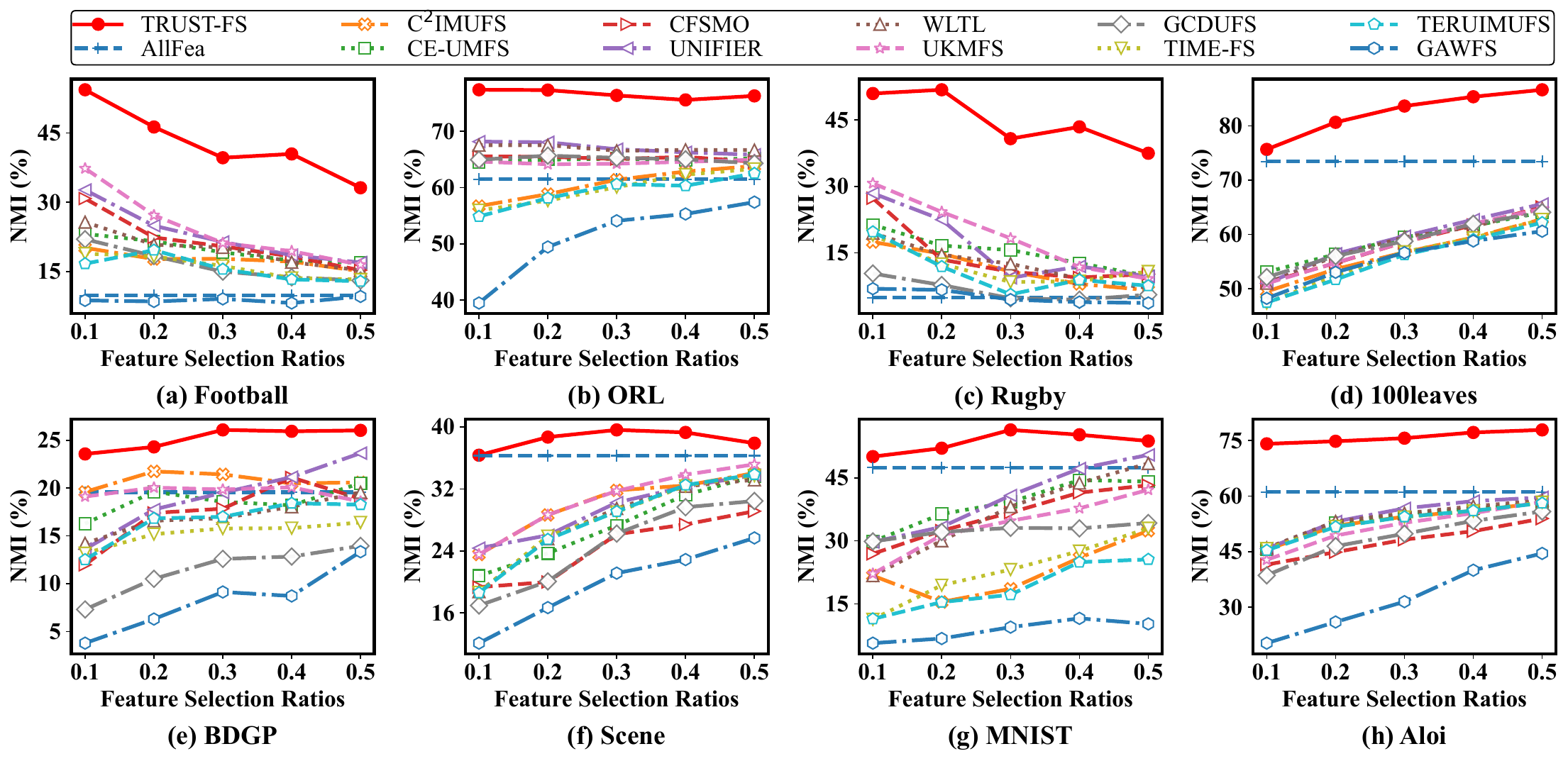}
	\caption{NMI of different methods on eight datasets under different feature selection ratios.}\label{NMI_feature_varying}
\end{figure*}

\begin{figure*}[t]
	\centering
	\includegraphics[width=\textwidth]{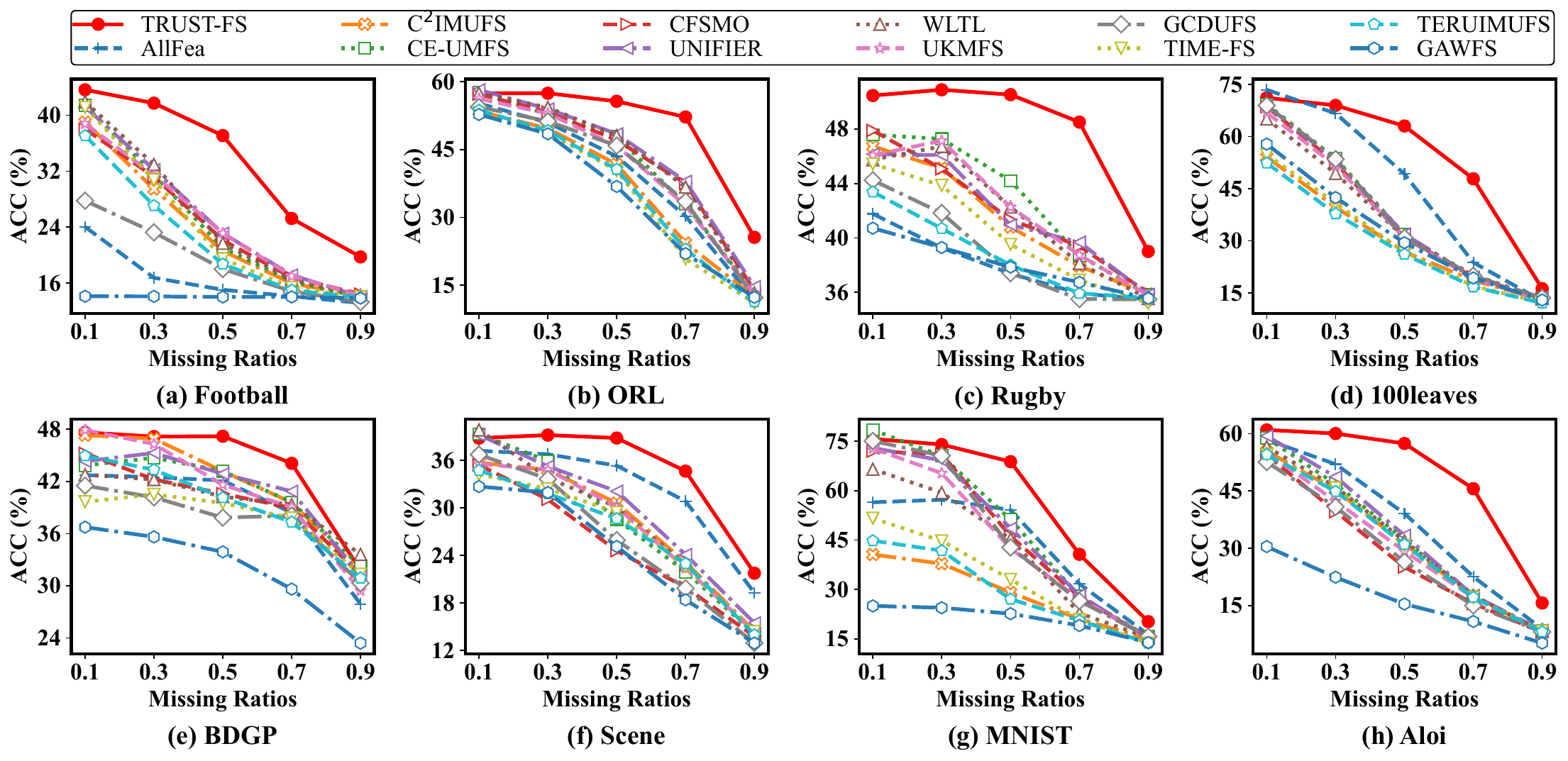}
	\caption{ACC of different methods on eight datasets with different missing ratios.}\label{ACC_missing_ratio_varying}
\end{figure*}

\begin{figure*}[!htbp]
	\centering
	\includegraphics[width=\textwidth]{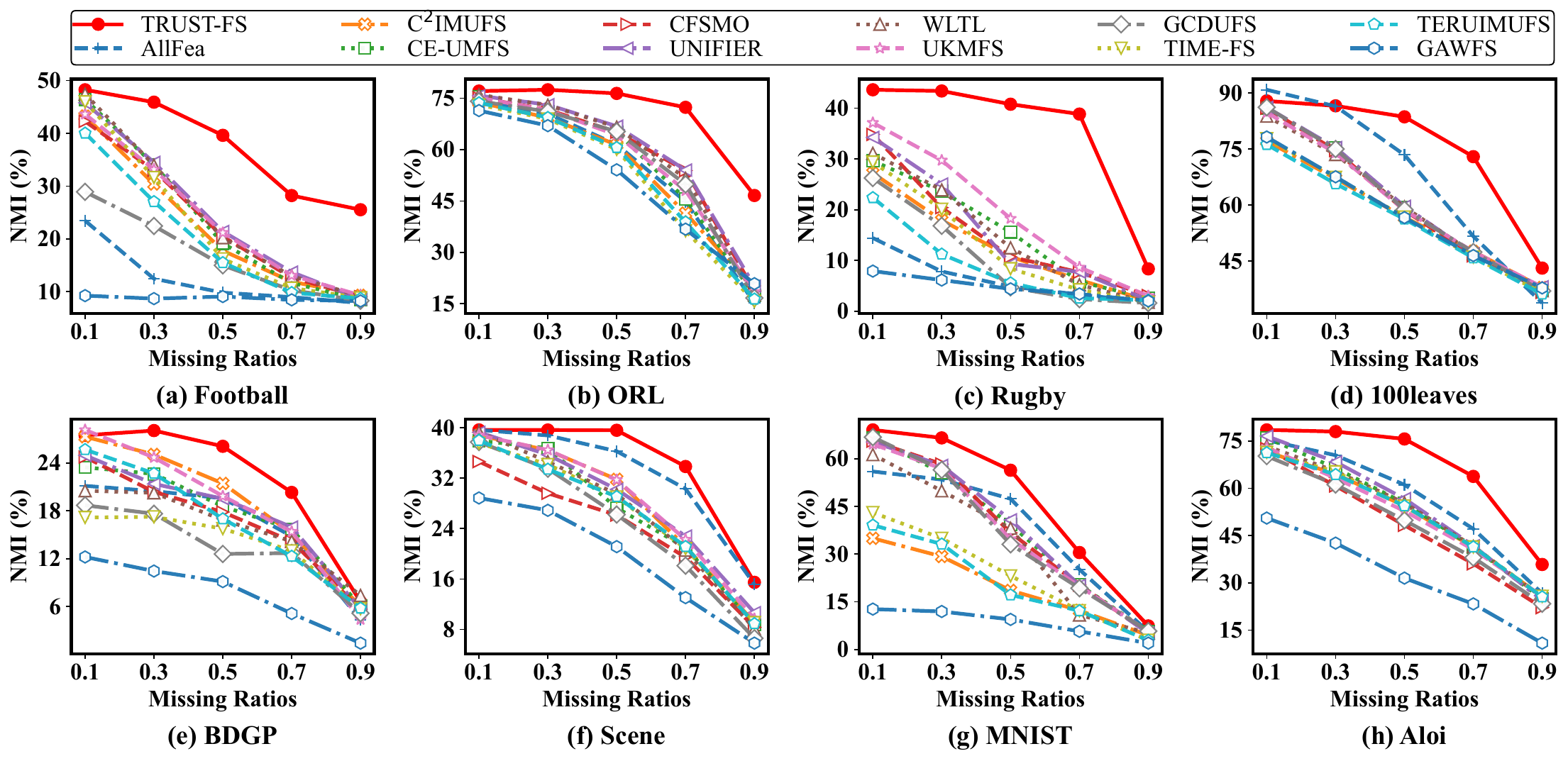}
	\caption{NMI of different methods on eight datasets with different missing ratios.}\label{NMI_missing_ratio_varying}
\end{figure*}

\subsubsection{Compared Methods}

To demonstrate the effectiveness of the proposed TRUST-FS, we compare it against several state-of-the-art (SOTA) methods, including four view-missing multi-view approaches (C$^2$IMUFS, UNIFIER, TERUIMUFS, and TIME-FS), five complete multi-view approaches (CE-UMFS, CFSMO, WLTL, UKMFS, and GCDUFS), and one single-view method (GAWFS). Brief introductions to these comparison methods are provided below:

$\bullet$ \textbf{AllFea} uses all features for comparison.

$\bullet$ \textbf{C$^2$IMUFS}\cite{C2IMUFS} combines adaptive view-weighted NMF with similarity reconstruction and consensus clustering to enable incomplete multi-view feature selection.

$\bullet$ \textbf{CE-UMFS}\cite{CE-UMFS} applies multi-view matrix factorization with HSIC and nuclear norm regularization to capture both exclusive and consistent information across views.

$\bullet$ \textbf{CFSMO}\cite{CFSMO} constructs graphs by integrating multi-order neighbor information and learns a shared latent representation for multi-view feature selection.

$\bullet$ \textbf{UNIFIER}\cite{UNIFIER} jointly learns similarity-induced graphs from both sample and feature spaces to facilitate adaptive view recovery and feature selection.

$\bullet$ \textbf{WLTL}\cite{WLTL} integrates multi-view spectral clustering with weighted low-rank tensor decomposition to generate high-quality pseudo labels for feature selection.

$\bullet$ \textbf{UKMFS}\cite{UKMFS} incorporates kernel-based graph fusion, low-rank constraints, and unsupervised hashing to select robust features across multiple views.

$\bullet$ \textbf{GCDUFS}\cite{GCDUFS} employs graph regularization, kernel dependency, and symmetric NMF to capture both consistent and complementary information from multiple views.

$\bullet$ \textbf{TIME-FS}\cite{TIME-FS} learns a consensus anchor graph and view-preference weights via CP decomposition to capture shared and view-specific information for missing-view recovery and feature selection.

$\bullet$ \textbf{TERUIMUFS}\cite{TERUIMUFS}  integrates self-representation learning with tensor low-rank constraints and sample diversity learning to perform feature selection.

$\bullet$ \textbf{GAWFS}\cite{GAWFS} uses non-negative matrix factorization and adaptive graph learning for data clustering, and identifies the most discriminative features through a feature weighting matrix.

Since some comparison methods require complete multi-view data and cannot be directly applied to incomplete datasets, we first impute missing values within each view using feature-wise mean imputation and then apply these methods. For a fair comparison, all methods being compared are tuned using grid search to achieve optimal performance. In our proposed method, we search for $\gamma$ in $\{2, 3, 4, 5, 6, 7\}$, and $\lambda$ and $\tau$  in $\{10^{-3}, 10^{-2}, 10^{-1}, 1, 10^{1}, 10^{2}, 10^{3}\}$. As determining the optimal number of selected features for a given dataset remains challenging~\cite{MFS-review,C2IMUFS}, we set the proportion of selected features to range from 10\% to 50\% in increments of 10\%. Then, $K$-means clustering is performed 20 times independently on the selected features, and the average results together with the standard deviations are reported.

\subsection{Performance Comparison}

To comprehensively evaluate the effectiveness of the proposed TRUST-FS, we report the clustering performance of all compared methods on eight datasets with varying feature selection and missing data ratios. Tables~\ref{table2} and~\ref{Table3} summarize the clustering results in terms of ACC and NMI, respectively, when the feature selection ratio is fixed at 30\% and the missing data ratio is set to 50\%.  The highest performance is shown in bold, and the second-best is underlined. In addition, a Wilcoxon signed-rank test is performed to determine whether the performance improvement of TRUST-FS over other methods is statistically significant. In the tables, the symbols $\bullet$ and $\circ$ indicate that TRUST-FS performs significantly better or worse than the compared methods at the 0.05 significance level, respectively.

Tables~\ref{table2} and~\ref{Table3} show that TRUST-FS consistently achieves superior performance compared to all competing methods across the eight benchmark datasets in terms of both ACC and NMI. As to Aloi dataset, TRUST-FS achieves substantial improvements of more than 14\% in ACC and 18\% in NMI compared to the second-best method. For Rugby dataset,  TRUST-FS shows increases of over 6\% in ACC and 22\% in NMI compared to the second-best method. Furthermore, TRUST-FS achieves improvements of more than 10\% in both metrics ACC and NMI on the Football and 100leaves datasets over the second-best method. On MNIST and ORL datasets, TRUST-FS achieves improvements of more than 7\% in ACC and 8\% in NMI, respectively, compared to the second-best method. TRUST-FS continues to achieve improvements of more than 3\% in both ACC and NMI on Scene and BDGP datasets relative to the second-best method. In addition, TRUST-FS consistently outperforms the AllFea baseline, demonstrating its capability to select a compact and representative subset of features while preserving overall effectiveness. Furthermore, across all datasets, TRUST-FS achieves the highest performance compared to recent view-missing approaches (i.e., C$^2$IMUFS, UNIFIER, TIME-FS, and TERUIMUFS), demonstrating its effectiveness in addressing the more general variable-missing scenario.

Since determining the optimal number of selected features for each dataset is often challenging, we further investigate how clustering performance varies across different feature selection ratios. Figs.~\ref{ACC_feature_varying} and~\ref{NMI_feature_varying} present the ACC and NMI results, respectively, for all methods as the feature selection ratio varies from 10\% to 50\%, with the missing ratio fixed at 50\%. As shown in these figures, TRUST-FS consistently outperforms all other methods across all eight datasets in terms of both ACC and NMI. To further illustrate the performance of TRUST-FS under varying missing ratios, Figs.~\ref{ACC_missing_ratio_varying} and~\ref{NMI_missing_ratio_varying} show the clustering results when the feature selection ratio is fixed at 30\%. As observed, TRUST-FS still outperforms other competing methods in most cases. The superior performance of TRUST-FS can be attributed to the integration of multi-view feature selection, missing value imputation, and reliable complementary similarity graph learning within a unified learning framework,  in which each component mutually reinforces the others.

\subsection{Visualization Analysis} 

\begin{figure}[!thpb]
	\centering
	\includegraphics[width=\columnwidth]{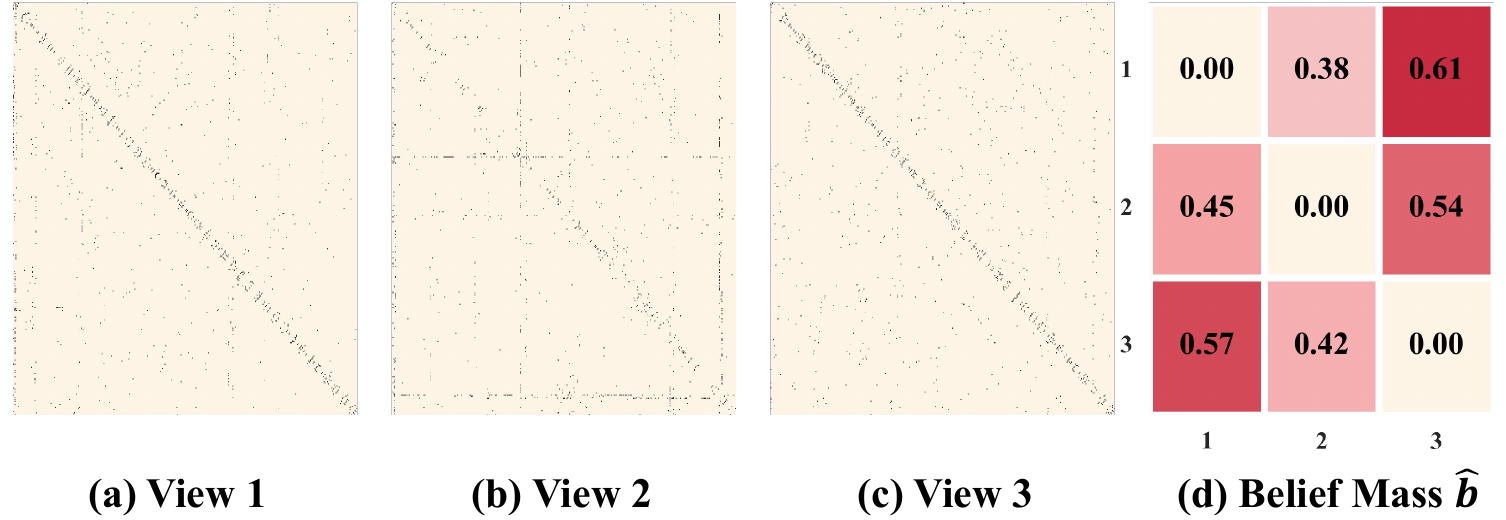} 
	\caption{Visualization of the similarity graph of each view and the belief mass \(\bm{\hat{b}}\) on ORL dataset.}
	\label{visual1}
\end{figure}

To intuitively demonstrate that the learned belief mass  can accurately reflect  the similarity among inter-view similarity graphs, we present visualizations of both the similarity graphs for each view and the corresponding belief mass obtained by TRUST-FS on ORL dataset with a missing ratio of 50\%, as shown in Fig.~\ref{visual1}. From this figure, we can see that the similarity graph of view 1 is more similar to that of view 3 than to that of view 2. The belief mass presented in Fig.~\ref{visual1}~(d) accurately reflects this relationship. This indicates that the proposed TRUST-FS can leverage more trustworthy information from other views to learn a more reliable similarity graph, even when a relatively high proportion of data is missing, thereby ultimately benefiting feature selection.

\subsection{Ablation Study} 

To demonstrate the effectiveness of the proposed module in TRUST-FS, we conduct ablation experiments by comparing it against three of its variants. Specifically: (i) TRUST-FS-I removes the adaptive imputation module from Eq.~(\ref{final_obj}), with missing entries filled using the feature-wise mean within each view; (ii) TRUST-FS-II replaces the A-WCP with standard CP decomposition; and (iii) TRUST-FS-III omits the reliable complementary graph learning module and instead constructs each view’s similarity graph using the $k$-nearest neighbor graph described in \cite{GMC}. 

\begin{figure}[!htbp]
	\centering
	\includegraphics[width=\columnwidth]{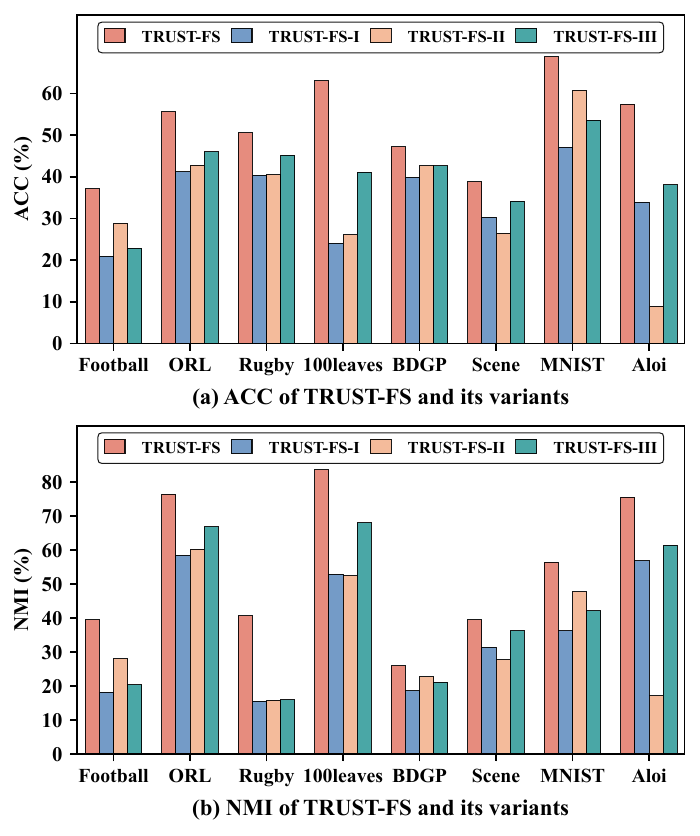} 
	\caption{Performance comparison of TRUST-FS and its three variants on eight datasets in terms of ACC and NMI.}%
	\label{ablation}
\end{figure}

Fig.~\ref{ablation} shows the ACC and NMI results of TRUST-FS and its three variants on eight datasets with a missing ratio of 50\% and 30\% of all features selected. As shown, TRUST-FS consistently outperforms all variants, confirming the effectiveness of joint learning of adaptive imputation and feature selection, A-WCP, and reliable complementary graph learning.
\subsection{Parameter Sensitivity and Convergence Analysis} 
\begin{figure}[!htbp]
	\centering
	\includegraphics[width=\columnwidth]{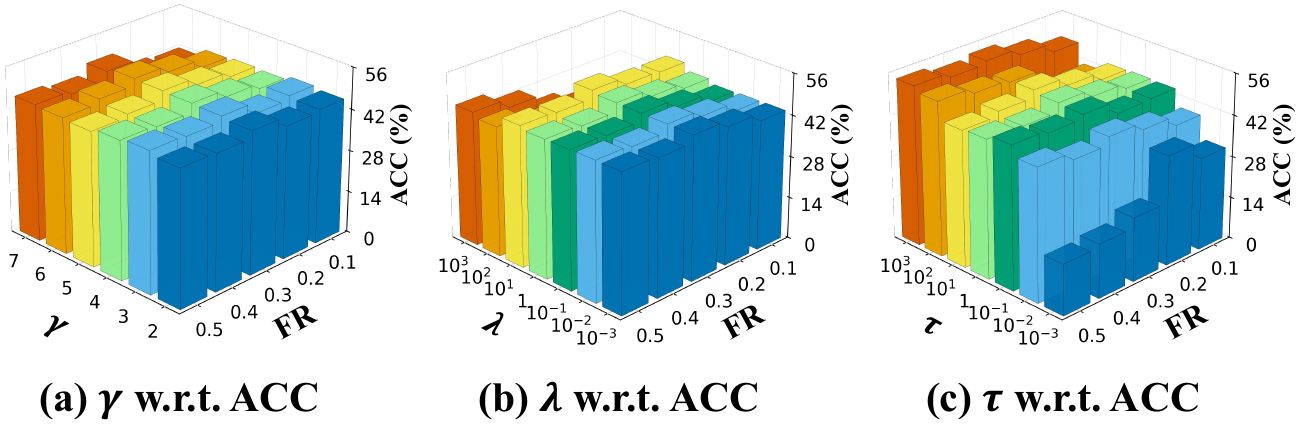} 
	\caption{ACC of TRUST-FS with varying parameters $\gamma$, $\lambda$, $\tau$ and feature selection ratios on ORL dataset.}
	\label{sensitivity}
\end{figure}

\begin{figure}[!htbp]
	\centering
	\includegraphics[width=\columnwidth]{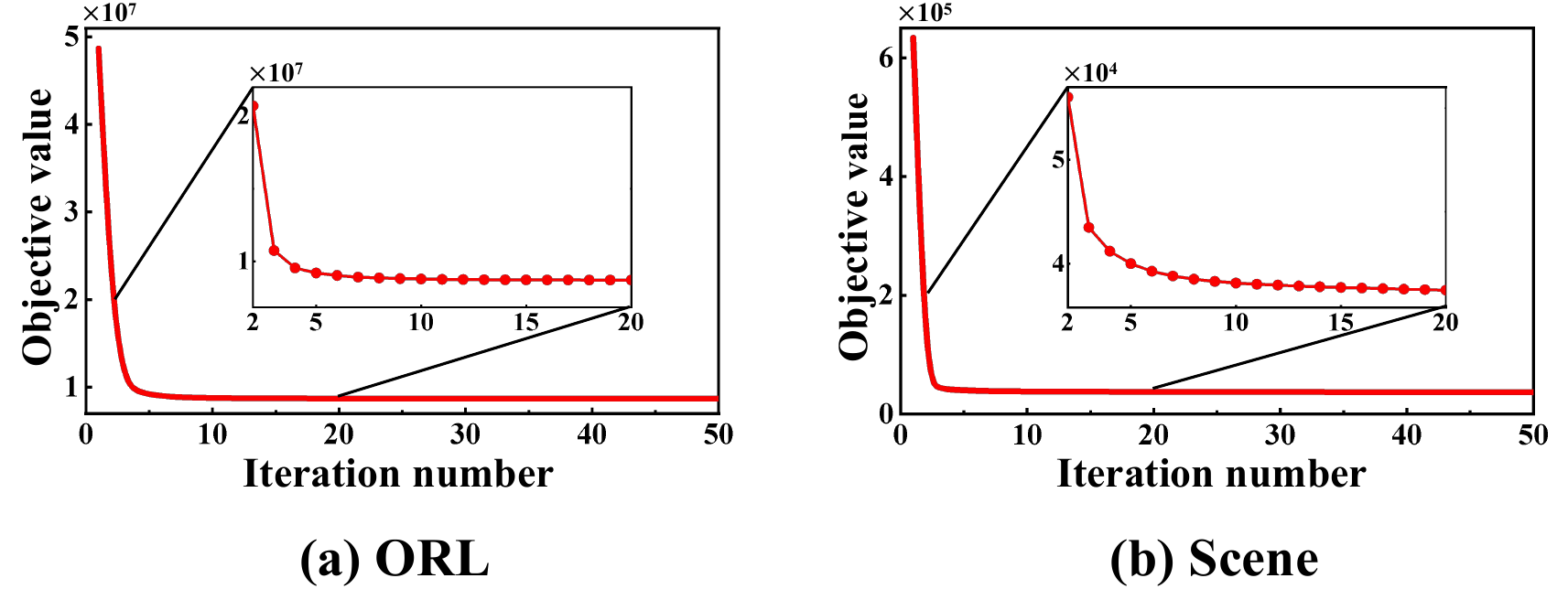} 
	\caption{Convergence curves of TRUST-FS on ORL and Scene datasets.}
	\label{convergence}
\end{figure}

We examine the sensitivity of TRUST-FS to the parameters \(\gamma\), \(\lambda\), and \(\tau\) by investigating how its performance varies with changes in these parameters and the feature selection ratio (FR). As shown in Fig.~\ref{sensitivity}, the ACC values exhibit only slight fluctuations when varying \(\gamma\) and \(\lambda\), whereas varying \(\tau\) results in relatively larger fluctuations.  Furthermore, Fig.~\ref{convergence} presents the convergence curves of TRUST-FS on ORL and Scene datasets, showing a sharp drop within the first few iterations and stabilization after approximately 20 iterations.

\section{Conclusion}
In this paper, we propose a novel MUFS method, TRUST-FS, for incomplete multi-view data with missing variables. Different from existing methods, we use the proposed A-WCP decomposition to integrate multi-view feature selection, missing variable imputation, and view weight learning into a unified tensor factorization framework. Meanwhile, Subjective Logic is employed to obtain trustworthy cross-view information, which is then used to guide similarity graph learning and improve their reliability in each view. Extensive experiments demonstrate the superiority of TRUST-FS over SOTA methods. In dynamic and open environments, the continuous evolution of data may cause previously important features to become less relevant. Therefore, in our future work, we will focus on developing incremental feature selection methods for multi-view datasets with missing variables.

\bibliographystyle{IEEEtran}  
\bibliography{citations}  

@article{ALOI,
	title={Multiclass from Binary: Expanding One-Versus-All, One-Versus-One and Ecoc-Based Approaches},
	author={Rocha, Anderson and Goldenstein, Siome Klein},
	journal={IEEE Transactions on Neural Networks and Learning Systems},
	volume={25},
	number={2},
	pages={289--302},
	year={2013},
}

@ARTICLE{100leaves,
	author={Yang, Jie and Lin, Chin-Teng},
	journal={IEEE Transactions on Emerging Topics in Computational Intelligence}, 
	title={Multi-View Adjacency-Constrained Hierarchical Clustering}, 
	year={2023},
	volume={7},
	number={4},
	pages={1126-1138},

}

@article{BDGP,
	author = {Cai, Xiao and Wang, Hua and Huang, Heng and Ding, Chris },
	title = {Joint stage recognition and anatomical annotation of drosophila gene expression patterns},
	journal = {Bioinformatics},
	volume = {28},
	number = {12},
	pages = {i16-i24},
	year = {2012},
}

@article{bestmap,
	title={Cluster Ensembles---A Knowledge Reuse Framework for Combining Multiple Partitions},
	author={Strehl, Alexander and Ghosh, Joydeep},
	journal={Journal of machine learning research},
	volume={3},
	pages={583--617},
	year={2002}
}

@ARTICLE{mnist,
	author={Deng, Li},
	journal={IEEE Signal Processing Magazine}, 
	title={The MNIST Database of Handwritten Digit Images for Machine Learning Research [Best of the Web]}, 
	year={2012},
	volume={29},
	number={6},
	pages={141-142}}

@inproceedings{rugby,
	title={Community Detection Based on Co-Regularized Nonnegative Matrix Tri-Factorization in Multi-View Social Networks},
	author={Yang, Longqi and Zhang, Liangliang and Pan, Zhisong and Hu, Guyu and Zhang, Yanyan},
	booktitle={Proceedings of the 5th IEEE International Conference on Big Data and Smart Computing},
	pages={98--105},
	year={2018},
}

@INPROCEEDINGS{Scene,
	author={Li, Fei-Fei and Pietro, Perona},
	booktitle={Proceedings of the 18th IEEE Computer Society Conference on Computer Vision and Pattern Recognition}, 
	title={A Bayesian Hierarchical Model for Learning Natural Scene Categories}, 
	year={2005},
	pages={524--531},
}

@article{FS-review,
	title={Feature Selection: A Data Perspective},
	author={Li, Jundong and Cheng, Kewei and Wang, Suhang and Morstatter, Fred and Trevino, Robert P and Tang, Jiliang and Liu, Huan},
	journal = {ACM Computing Surveys},
	volume={50},
	number={6},
	pages={1--45},
	year={2017},
}

@article{multi-view-survey-2,
	title={A Survey of Multi-View Machine Learning},
	author={Sun, Shiliang},
	journal={Neural Computing and Applications},
	volume={23},
	number={7},
	pages={2031--2038},
	year={2013},
	
}

@inproceedings{missing-value1,
	title={Active Vision for Early Recognition of Human Actions},
	author={Wang, Boyu and Huang, Lihan and Hoai, Minh},
	booktitle={Proceedings of the 33rd IEEE Computer Society Conference on Computer Vision and Pattern Recognition},
	pages={1081--1091},
	year={2020}
}

@article{dirichlet1,
	title={Concepts of Independence for Proportions with a Generalization of the Dirichlet Distribution},
	author={Connor, Robert J and Mosimann, James E},
	journal={Journal of the American Statistical Association},
	volume={64},
	number={325},
	pages={194--206},
	year={1969},
}

@article{dirichlet2,
	title={A Characterization of the Dirichlet Distribution},
	author={Darroch, JN and Ratcliff, D},
	journal={Journal of the American Statistical Association},
	volume={66},
	number={335},
	pages={641--643},
	year={1971},
}

@inproceedings{zhang2024efficient ,
	title={Efficient Multi-View Unsupervised Feature Selection with Adaptive Structure Learning and Inference},
	author={Zhang, Chenglong and Fang, Yang and Liang,  Xinyan and Zhang, Han and  Zhou, Peng and Wu, Xingyu and Yang,  Jie  and Jiang, Bingbing and Sheng, Weiguo},
	booktitle={Proceedings of the 33rd International Joint Conference on Artificial Intelligence},
	pages={5443--5452},
	year={2024}
}

@inproceedings{Zhang2024Scalable,
	author = {Zhang, Chenglong and Liang, Xinyan and Zhou, Peng and Ling, Zhaolong and Zhang, Yingwei and Wu, Xingyu and Sheng, Weiguo and Jiang, Bingbing},
	title = {Scalable Multi-View Unsupervised Feature Selection with Structure Learning and Fusion},
	year = {2024},
	booktitle = {Proceedings of the 32nd ACM International Conference on Multimedia},
	pages = {5479–5488},
}

@article{multi-view-learning,
	title={Multi-View Learning Overview: Recent Progress and New Challenges},
	author={Zhao, Jing and Xie, Xijiong and Xu, Xin and Sun, Shiliang},
	journal={Information Fusion},
	volume={38},
	pages={43--54},
	year={2017},
}

@article{unlabeled,
	title={Advanced Unsupervised Learning: A Comprehensive Overview of Multi-View Clustering Techniques},
	author={Moujahid, Abdelmalik and Dornaika, Fadi},
	journal={Artificial Intelligence Review},
	volume={58},
	number={8},
	pages={1--52},
	year={2025},
}

@article{MFS-review,
	title={Feature Selection with Multi-View Data: A Survey},
	author={Zhang, Rui and Nie, Feiping and Li, Xuelong and Wei, Xian},
	journal={Information Fusion},
	volume={50},
	pages={158--167},
	year={2019},
}

@article{missing-variable2,
	title={Incomplete Multi-View Learning: Review, Analysis, and Prospects},
	author={Tang, Jingjing and Yi, Qingqing and Fu, Saiji and Tian, Yingjie},
	journal={Applied Soft Computing},
	volume={153},
	pages={111278},
	year={2024},
}

@inproceedings{NDFS,
	title={Unsupervised feature selection using nonnegative spectral analysis},
	author={Li, Zechao and Yang, Yi and Liu, Jing and Zhou, Xiaofang and Lu, Hanqing},
	booktitle={Proceedings of the AAAI conference on artificial intelligence},
	volume={26},
	number={1},
	pages={1026--1032},
	year={2012}
}

@inproceedings{LS,
	title={Laplacian Score for Feature Selection},
	author={He, Xiaofei and Cai, Deng and Niyogi, Partha},
	booktitle={Proceedings of the 18th International Conference on Neural Information Processing Systems},
	pages={507--514},
	year={2005}
}

@article{GAWFS,
	title={A General Adaptive Unsupervised Feature Selection with Auto-Weighting},
	author={Liao, Huming and Chen, Hongmei and Yin, Tengyu and Yuan, Zhong and Horng, Shi-Jinn and Li, Tianrui},
	journal={Neural Networks},
	volume={181},
	pages={106840},
	year={2025}
}

@book{SL1,
	title={Subjective Logic: A Formalism for Reasoning Under Uncertainty},
	author={Jsang, Audun},
	year={2018},
	publisher={Springer}
}

@article{SL2,
	title={What Uncertainties Do We Need in Bayesian Deep Learning for Computer Vision?},
	author={Kendall, Alex and Gal, Yarin},
	journal={Advances in Neural Information Processing Systems},
	volume={30},
	year={2017},
	pages = {5574--5584}
}

@inproceedings{UNIFIER,
	title={Unified View Imputation and Feature Selection Learning for Incomplete Multi-View Data},
	author={Huang, Yanyong and Shen, Zongxin and Li, Tianrui and Lv, Fengmao},
	booktitle={Proceedings of the 33rd International Joint Conference on Artificial Intelligence},
	pages={4192--4200},
	year={2024}
}

@article{TERUIMUFS,
	title={Tensor-based unsupervised feature selection for error-robust handling of unbalanced incomplete multi-view data},
	author={Yang, Xuanhao and Che, Hangjun and Leung, Man-Fai},
	journal={Information Fusion},
	volume={114},
	pages={102693},
	year={2025},
}

@inproceedings{TIME-FS,
	title={TIME-FS: Joint Learning of Tensorial Incomplete Multi-View Unsupervised Feature Selection and Missing-View Imputation},
	author={Huang, Yanyong and Lu, Minghui and Huang, Wei and Yi, Xiuwen and Li, Tianrui},
	booktitle={Proceedings of the 39th AAAI Conference on Artificial Intelligence},
	pages={17503--17510},
	year={2025}
}

@article{CFSMO,
	title={Multi-View Unsupervised Complementary Feature Selection with Multi-Order Similarity Learning},
	author={Cao, Zhiwen and Xie, Xijiong},
	journal={Knowledge-Based Systems},
	volume={283},
	pages={111172},
	year={2024},
}

@article{SCMvFS,
	title={Structure Learning with Consensus Label Information for Multi-View Unsupervised Feature Selection},
	author={Cao, Zhiwen and Xie, Xijiong},
	journal={Expert Systems with Applications},
	volume={238},
	pages={121893},
	year={2024},
}

@article{C2IMUFS,
	author={Huang, Yanyong and Shen, Zongxin and Cai, Yuxin and Yi, Xiuwen and Wang, Dongjie and Lv, Fengmao and Li, Tianrui},
	journal={IEEE Transactions on Knowledge and Data Engineering}, 
	title={C$^2$IMUFS: Complementary and Consensus Learning-Based Incomplete Multi-View Unsupervised Feature Selection}, 
	year={2023},
	volume={35},
	number={10},
	pages={10681-10694},
}

@inproceedings{UKMFS,
	title={Unsupervised Kernel-Based Multi-View Feature Selection with Robust Self-Representation and Binary Hashing},
	author={Hu, Rongyao and Gan, Jiangzhang and Zhan, Mengmeng and Li, Li and Wei, Mengling},
	booktitle={Proceedings of the 39th AAAI Conference on Artificial Intelligence},
	pages={17287--17294},
	year={2025}
}

@article{WLTL,
	title={Unsupervised Multi-View Feature Selection Based on Weighted Low-Rank Tensor Learning and Its Application in Multi-Omics Datasets},
	author={Wang, Daoyuan and Wang, Lianzhi and Chen, Wenlan and Wang, Hong and Liang, Cheng},
	journal={Engineering Applications of Artificial Intelligence},
	volume={143},
	pages={110041},
	year={2025},
}

@article{CE-UMFS,
	title={Consistency--Exclusivity Guided Unsupervised Multi-View Feature Selection},
	author={Zhou, Shixuan and Song, Peng},
	journal={Neurocomputing},
	volume={569},
	pages={127119},
	year={2024},
}

@article{GCDUFS,
	title={Graph--Regularized Consensus Learning and Diversity Representation for Unsupervised Multi-View Feature Selection},
	author={Xu, Shengke and Xie, Xijiong and Cao, Zhiwen},
	journal={Knowledge-Based Systems},
	pages={113043},
	year={2025},
}

@article{KKT,
	title={Karush-Kuhn-Tucker Conditions},
	author={Gordon, Geoff and Tibshirani, Ryan},
	journal={Optimization},
	volume={10},
	number={725/36},
	pages={725},
	year={2012}
}

@article{CP,
	title={Tensor Decompositions and Applications},
	author={Kolda, Tamara G and Bader, Brett W},
	journal = {SIAM Review},
	volume={51},
	number={3},
	pages={455--500},
	year={2009},
}

@article{CP-D,
	title={Analysis of Individual Differences in Multidimensional Scaling via an N-Way Generalization of “Eckart-Young” Decomposition},
	author={Carroll, J Douglas and Chang, Jih-Jie},
	journal={Psychometrika},
	volume={35},
	number={3},
	pages={283--319},
	year={1970},
}

@article{GMC,
	title={GMC: Graph-Based Multi-View Clustering},
	author={Wang, Hao and Yang, Yan and Liu, Bing},
	journal={IEEE Transactions on Knowledge and Data Engineering}, 
	volume={32},
	number={6},
	pages={1116--1129},
	year={2019},
}

@inproceedings{l21,
	author = {Nie, Feiping and Huang, Heng and Cai, Xiao and Ding, Chris},
	booktitle = {Proceedings of the 23rd International Conference on Neural Information Processing Systems},
	title = {Efficient and Robust Feature Selection via Joint $\ell$2,1-Norms Minimization},
	pages = {1813--1821},
	year = {2010}
}

@article{simulate-missing-variable,
	title={Joint Embedding Learning and Low-Rank Approximation: A Framework for Incomplete Multiview Learning},
	author={Tao, Hong and Hou, Chenping and Yi, Dongyun and Zhu, Jubo and Hu, Dewen},
	journal={IEEE Transactions on Cybernetics},
	volume={51},
	number={3},
	pages={1690--1703},
	year={2019},
}

@ARTICLE{convergence,
	author={Ding, Chris H.Q. and Li, Tao and Jordan, Michael I.},
	journal={IEEE Transactions on Pattern Analysis and Machine Intelligence}, 
	title={Convex and Semi-Nonnegative Matrix Factorizations}, 
	year={2010},
	volume={32},
	number={1},
	pages={45-55},
}

\vfill

\end{document}